\pdfoutput=1

\documentclass[11pt]{article}

\usepackage{acl}

\usepackage{times}
\usepackage{latexsym}
\usepackage{bbm} 
\usepackage{amssymb}
\usepackage[T1]{fontenc}
\usepackage{geometry}
\usepackage{CJK}

\usepackage[utf8]{inputenc}

\usepackage{microtype}
\usepackage{amsmath}

\usepackage{inconsolata}

\usepackage{graphicx}

\usepackage{subcaption}
\usepackage{etoolbox}
\usepackage{mdframed}
\usepackage{makecell}
\usepackage{amsfonts}

\usepackage{booktabs}
\usepackage{geometry}
\usepackage{etoolbox}
\usepackage{xcolor}
\usepackage{multirow}
\usepackage{booktabs}
\usepackage{afterpage}

\title{Let the Expert Stick to His Last: Expert-Specialized Fine-Tuning for Sparse Architectural Large Language Models}

\author{Zihan Wang$^{12}$\thanks{Work done during internship at DeepSeek.}, Deli Chen$^{1}$, Damai Dai$^{1}$, Runxin Xu$^{1}$, Zhuoshu Li$^{1}$, Y. Wu$^{1}$ \\
$^{1}$DeepSeek AI \\
$^{2}$Northwestern University\\
\{zw, victorchen\}@deepseek.com}

\begin{document}
\maketitle

\begin{abstract}
Parameter-efficient fine-tuning~(\textbf{PEFT}) is crucial for customizing Large Language Models (\textbf{LLMs}) with constrained resources.
Although there have been various PEFT methods for dense-architecture LLMs, PEFT for sparse-architecture LLMs is still underexplored.
In this work, we study the PEFT method for LLMs with the Mixture-of-Experts (\textbf{MoE}) architecture and the contents of this work are mainly threefold:
(1) We investigate the dispersion degree of the activated experts in customized tasks, and found that the routing distribution for a specific task tends to be highly concentrated, while the distribution of activated experts varies significantly across different tasks.
(2) We propose \textbf{E}xpert-\textbf{S}pecialized \textbf{F}ine-\textbf{T}uning, or ESFT, which tunes the experts most relevant to downstream tasks while freezing the other experts and modules; experimental results demonstrate that our method not only improves the tuning efficiency, but also matches or even surpasses the performance of full-parameter fine-tuning.
(3) We further analyze the impact of the MoE architecture on expert-specialized fine-tuning. We find that MoE models with finer-grained experts are more advantageous in selecting the combination of experts that are most relevant to downstream tasks, thereby enhancing both the training efficiency and effectiveness. Our code is available at \url{https://github.com/deepseek-ai/ESFT}.

%
\end{abstract}

\section{Introduction}

As the parameter scale of large language models~(\textbf{LLMs}) continues to increase~\citep{Model_LLAMA3, Model_mixtral8x22b, Model_DeepSeekV2, Model_qwen1_5}, parameter-efficient fine-tuning~(\textbf{PEFT}) methods \citep{Peft_Survey} are becoming increasingly important in adapting pre-trained LLMs to downstream customization tasks.
However, existing works on PEFT like low-rank adaptation (LoRA) and P-Tuning~\citep{5-LoRA, 47-P-tuning-v2} have primarily focused on dense-architecture LLMs, with research on sparse-architecture LLMs still being markedly insufficient.

In this work, we focus on exploring PEFT techniques within the Mixture-of-Experts (\textbf{MoE}) LLMs~\citep{Model_mixtral_8x7b, Model_DBRX}, as introduced in \S\ref{sec:moe}. 
Unlike dense models where all tasks are handled by the same parameters, in the MoE architecture, different tasks are processed by distinct activated experts~\citep{gshard, switch}.
Observations indicate that task specialization in expert systems is the key to the MoE LLM performance~\citep{DeepSeek_MoE}. 
We further illustrate such specialization in \S\ref{sec:specialization-in-moe} that experts activated by the same task's data are concentrated, while those for different tasks vary significantly, suggesting MoE models use specialized expert combinations to handle different tasks. 
Motivated by this, we propose Expert-Specialized Fine-Tuning~(\textbf{ESFT}), as illustrated in \S\ref{sec:esft}. 
ESFT only tunes the experts with the highest affinity to the task, while freezing the parameters of other experts and modules.

The primary advantages of ESFT lie in two aspects: 
(1)~\textbf{Maintaining Expert Specialization}: ESFT prevents the decrement of specialization in full-parameter fine-tuning, where experts not adept at the task also update their parameters. 
Experimental results in \S\ref{sec:mainresult} show that ESFT can achieve aligned or even superior performance in downstream tasks compared to full-parameter fine-tuning, and better maintains performance in general tasks.
(2)~\textbf{Saving Computation Resources}: ESFT only trains the parameters of the selected experts, which effectively reduces the storage of up to 90\% and training time up to 30\% compared to full-parameter fine-tuning, as shown in \S\ref{sec:efficiencyresults}.

Besides, we delve deeper into the working mechanism of the ESFT method. We analyze the expert selection process in \S\ref{sec:epselection} and demonstrate how ESFT leverages specialized experts effectively, as selecting 5-15\% experts can achieve promising performance in different tasks.
We investigate the efficiency of ESFT under different computational constraints in \S\ref{sec:param}, showcasing its ability to leverage training resources efficiently compared to other PEFT methods like LoRA. 
Our studies in \S\ref{sec:abl-shared} analyze the effects of shared and non-shared parameters in the model on specialized and general performance, pointing out the priority to selectively train non-shared parameters in ESFT.
Through ablation studies in \S\ref{sec:abl}, we highlight the importance of our expert relevance scores and the fine-grained expert segmentation architecture. 



\section{Related Work}

\subsection{Parameter-efficient fine-tuning for dense architectural LLMs}
\label{related_work:peft}
The goal of parameter-efficient fine-tuning~\citep{Peft_Survey} is to efficiently customize LLMs for downstream tasks, while existing studies primarily focus on dense architectural LLMs. 
PEFT methods for dense models can generally be categorized into three approaches:
(1)~\textbf{Adding new parameters}: methods of this kind fix the existing model parameters and fine-tune the model on a small number of newly added parameters. Adapter~\citep{9-peft-transfer-learning, 4-AdapterFusion, 14-unified-view-transfer-peft, 29-adamix} and Soft Prompt~\citep{11-prefix-tuning, 47-P-tuning-v2, zhang2023towards,13-prompt-tuning} are two typical representatives of this category of methods.
(2)~\textbf{Selecting existing parameters}: methods of this type fine-tune a limited part of existing parameters, while keeping the majority of the other parameters fixed. Based on whether the trainable parameter space is continuous, these methods can generally be divided into structured training~\citep{10-diff-purning, gheini2021cross, he2023sensitivity, 53-FAR} and unstructured training~\citep{liao2023parameter, ansell2021composable, 36-fish-masks, 54-raise-child}.
(3)~\textbf{Applying low-rank adaptation}: LoRA~\citep{5-LoRA,fomenko2024note} is a widely-used PEFT method, which decomposes the origin weight matrices into low-rank components. Subsequent works~\citep{adalora, SORA, lin2024lora,liu2023moelora} have introduced numerous improvements to the original LoRA method.
However, the study of PEFT in sparse models is still scarce. In this work, we select and tune part of the experts based on their downstream task affinity, as a unique selection dimension exclusive to the sparse MoE architecture.

\subsection{Coarse- and Fine-grained MoE LLMs}
\label{related_work:moe}
Compared to dense LLMs (e.g., LLaMA series, \citealp{Model_LLaMA1, Model_LLaMA2}), MoE LLMs (e.g., Mixtral series, \citealp{Model_mixtral8x22b, Model_mixtral_8x7b}) can increase model size while saving training and inference costs.
Based on the granularity of experts, existing large MoE models can generally be divided into two categories: coarse- and fine-grained expert LLMs. 
Most existing MoE LLMs~\citep{gshard,switch, hash, stablemoe, Model_JetMoE} have coarse-grained experts where the number of experts is very limited.
For example, 2 out of 8 experts are activated for Mixtral MoE series~\citep{Model_mixtral8x22b, Model_mixtral_8x7b} and Grok-V1~\citep{Model_Grok1}.
As a result, a single expert has to learn complicated patterns from different domain tasks simultaneously.
To address this issue, DeepSeek MoE~\citep{DeepSeek_MoE} has introduced fine-grained expert segmentation. In the DeepSeek-V2~\citep{Model_DeepSeekV2}, there are as many as 162 experts, with 8 active experts (8 out of 66 experts are activated for the DeepSeek-V2-Lite).
The fine-grained division of experts ensures a high degree of specialization among the experts. Moreover, the specialized expert system enables the selection of experts that are most relevant to the task for efficient tuning.

\section{Methods}

\begin{figure*}[h]
    \centering
    \includegraphics[width=\textwidth]{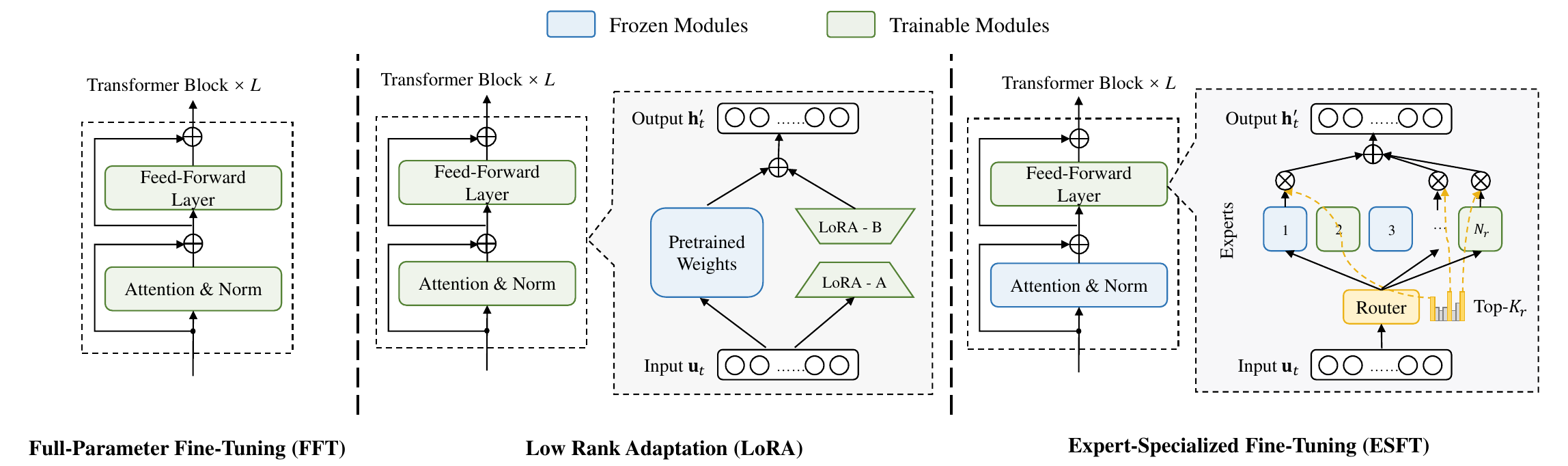}
    \caption{Comparison between Expert-Specialized Fine-Tuning (ESFT) and other fine-tuning methods. FFT trains all parameters. LoRA combines pre-trained weights with low-rank matrices to reduce training costs. ESFT only trains a subset of experts in a Mixture-of-Expert (MoE) architecture, optimizing efficiency and task specialization. }
    \label{fig:method}
\end{figure*}

\subsection{Preliminaries: Mixture-of-Experts for Transformers}
\label{sec:moe}
Mixture-of-Experts (MoE) for Transformers replace Feed-Forward Networks (FFNs) with MoE layers. Each MoE layer consists of multiple experts structurally identical to a FFN. Tokens are assigned to and processed by a subset of the most relevant experts based on their affinity scores, ensuring computational efficiency in MoE layers. The output hidden state $\mathbf{h}_t^l$ of the $t$-th token in the $l$-th MoE layer is computed as:
\begin{equation}
\mathbf{h}_t^l = \sum_{i=1}^{N} \left(g_{i,t} \text{FFN}^n_i(\mathbf{u}_t^l)\right) + \mathbf{u}_t^l,
\end{equation}
\begin{equation}
g_{i,t} = \begin{cases}
s_{i,t}, & s_{i,t}{\small\in}\text{TopK}(\{s_{j,t}|1 {\small \leqslant}  j {\small \leqslant}  N\}, K), \\
0, & \text{otherwise},
\end{cases}
\end{equation}
\begin{equation}
s_{i,t} = \text{Softmax}_i\left(\mathbf{u}_t^{l\top} \mathbf{e}_i^l\right), 
\end{equation}
where $N$ denotes the total number of experts, $\text{FFN}_i(\cdot)$ is the $i$-th expert FFN, $g_{i,t}$ denotes the gate value for the $i$-th expert, $s_{i,t}$ denotes the token-to-expert affinity, $\text{TopK}(\cdot, K)$ denotes the set comprising $K$ highest affinity scores among those calculated for the $t$-th token and all $N$ experts, and $\mathbf{e}_i^l$ is the centroid of the $i$-th expert in the $l$-th layer.

Recently, DeepSeekMoE~\citep{DeepSeek_MoE} proposes enhancements to the MoE architecture through several techniques, including
(1) Fine-grained segmentation, segmenting each expert into multiple smaller ones and keeping the same fraction of experts to process each token, allowing specialization in different knowledge types while maintaining the same computational cost. (2) Shared expert isolation, leveraging shared experts that process all tokens to capture common knowledge, reducing parameter redundancy and enhancing efficiency.
The output of an MoE layer in DeepSeekMoE is:
\begin{equation}
\mathbf{h}_t^l{\small=}\sum_{i=1}^{K_s} \text{FFN}^s_i(\mathbf{u}_t^l){\small+}\sum_{i=1}^{N} (g_{i,t} \text{FFN}^n_i{\small (}\mathbf{u}_t^l{\small )}){\small+}\mathbf{u}_t^l,
\end{equation}
\begin{equation}
g_{i,t} = \begin{cases}
s_{i,t},~s_{i,t}{\small\in}\text{TopK}(\{s_{j,t}|1 {\small \leqslant} j {\small \leqslant} N\}, K{\small -}K_s), \\
0,~~~\text{otherwise},
\end{cases}
\end{equation}
where $K_s$ is the number of shared experts, $\text{FFN}^s_i$ and $\text{FFN}^n_i$ denote the shared and non-shared experts, respectively. Each expert is segmented into $m$ ones, with $N$ and $K$ also multiplied by $m$ times compared to the coarse-grained architecture.


\subsection{Probing Task-Specific Expert \textit{Specialization} in MoE Models}
\label{sec:specialization-in-moe}
\begin{figure}[t]
    \centering
    \includegraphics[width=\linewidth]{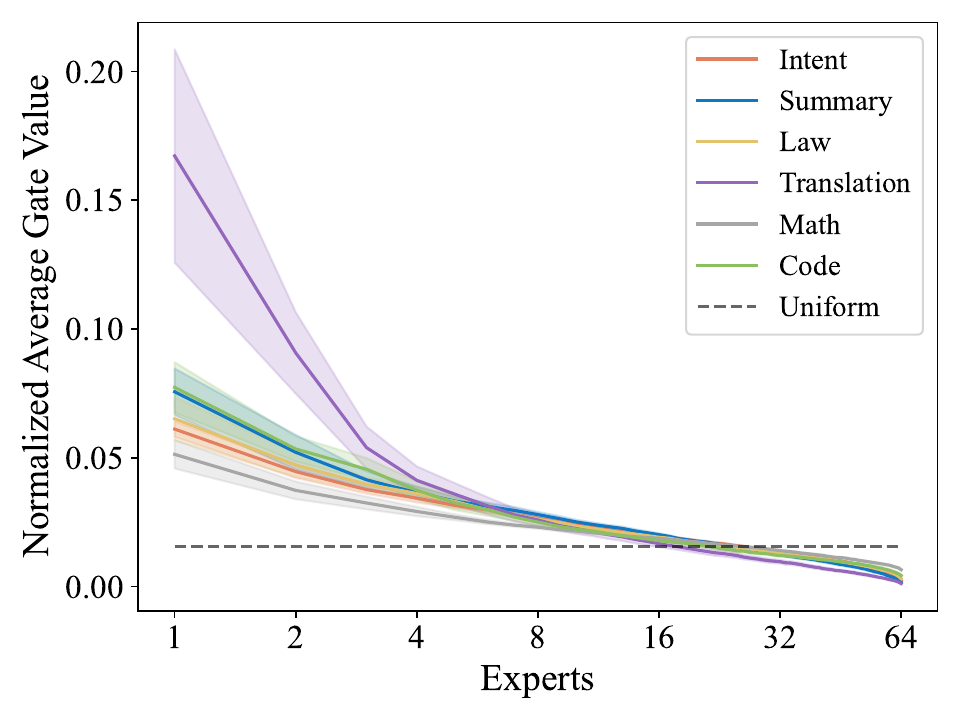}
    \caption{Top Expert distribution for specific tasks. Shaded areas represent variance across layers. The figure shows that few experts handle most gate values, highlighting expert specialization for different tasks. }
    \label{fig:expert_routing}
\end{figure}

Despite the significant success of MoE LLMs, a clear understanding of the underlying mechanism remains elusive.
We conduct probing experiments to understand how non-shared experts are utilized across various tasks. These tasks, as detailed in \S\ref{sec:benchmarks}, include general domains like math and code, as well as specialized domains like intent recognition, summarization, legal judgment prediction, and translation. These experiments reveal the expert specialization in MoE models in two aspects:

\paragraph{Expert Routing is Concentrated in the Same Task} We investigate the distribution of normalized gate values, i.e., the sum of all expert-token gate values for each expert, divided by the total across all experts.
Figure~\ref{fig:expert_routing} displays this distribution, where the experts are sorted by their normalized values from high to low.
The figure shows that a small subset of experts handles the majority of gate values, indicating the model's and concentrated expert allocation for a specific task.

\begin{figure}[t]
    \centering
    \includegraphics[width=1.05\linewidth]{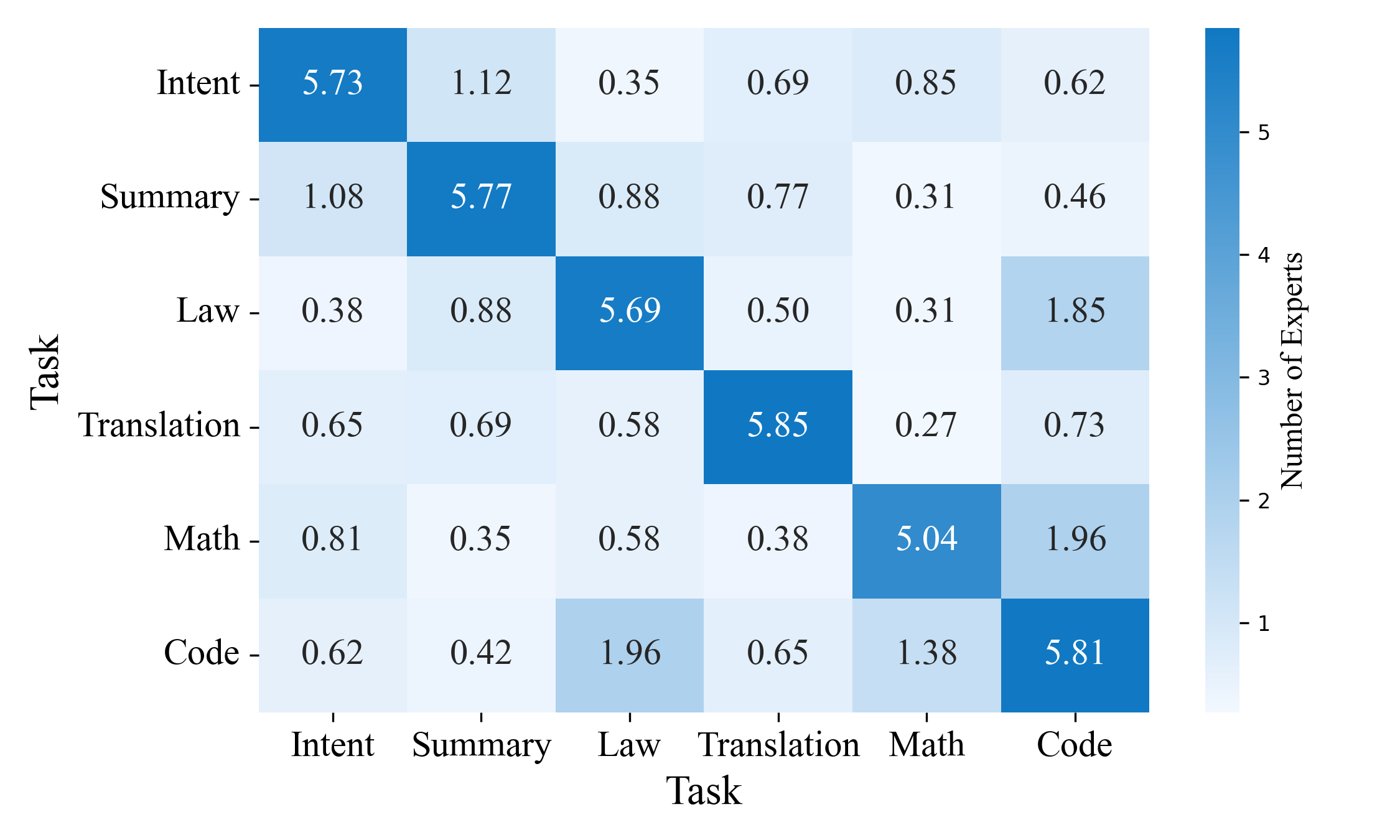}
    \caption{The average number of shared Top-6 routed experts across tasks. The values are averaged by layer, indicating that the sets of experts used for the same task are consistent while different tasks are distinct.}
    \label{fig:sharedexperts}
\end{figure}

\paragraph{Active Experts Vary Significantly across Tasks} 
We investigate the joint distribution of experts across tasks. 
Figure~\ref{fig:sharedexperts} shows a heatmap of the shared Top-6 experts for two independent data samples per task averaged across layers. This indicates the degree of overlap of experts used within the same task or between different tasks.
Off-diagonal values are near 0, and diagonal values are near 6, indicating that the same task uses similar experts, while different tasks use different sets. 

\subsection{Expert-Specialized Fine-tuning (ESFT)}

\label{sec:esft}
The highly specialized expert system suggests that different experts can be optimized for specific tasks. Inspired by this, we propose Expert-Specialized Fine-Tuning (ESFT) for MoE LLM customization, which selectively fine-tunes the most relevant experts for downstream tasks to enhance computational efficiency and maintain expert specialization. Figure~\ref{fig:method} illustrates the differences between our method and existing methods. Below, we introduce our method step by step.

\paragraph{Data Sampling}
We randomly sample a subset $D_s = \{(x_i, y_i)\}_{i=1}^{N_s}$ from the training data $D = \{(x_i, y_i)\}_{i=1}^{N}$ for expert selection, where $x_i$ and $y_i$ denote the input and label, respectively. Empirically, we find that a subset of $32$ concatenated samples, each with a fixed length of $L = 4096$, is robust enough to select the most relevant experts for a task. We detail this claim in Appendix~\ref{app:samplesize}.

\paragraph{Expert Relevance Score}
We propose two methods to calculate the relevance of an expert to a task based on its affinity to the sample tokens, defined as average gate score and token selection ratio, respectively. Both methods assess each expert's relevance to downstream tasks and can be chosen based on task-specific experimental performance.

\textbf{Average Gate Score (\textbf{ESFT-Gate})}~~This score calculates the average affinity of expert $e_i$ to all tokens in the sampled data. It is defined as:
\begin{equation}
g_i^l = \frac{1}{N_s} \sum_{j=1}^{N_s} \frac{1}{L_j} \sum_{k=1}^{L_j} g_{i,k}^l,
\end{equation}
where $L_j$ is the length of the input sequence $x_j$ in the sampled data $D_s$.

\textbf{Token Selection Ratio (\textbf{ESFT-Token})}~~This score calculates the ratio of tokens for which expert $e_i$ is selected. It is defined as:
\begin{equation}
r_i^l = \frac{1}{N_s} \sum_{j=1}^{N_s} \frac{1}{L_j}\sum_{k=1}^{L_j}\frac{\mathbbm{1}\left(g_{i,k}^l > 0\right)}{K},
\end{equation}
where $\mathbbm{1}\left(g_{i,k}^l > 0\right)$ is an indicator  that equals 1 if the gate score $g_{i,k}^l$ is positive, and 0 otherwise. $K$ is the number of experts selected per token. 

\paragraph{Expert Selection and Fine-tuning}
For each MoE layer $l$, we select a subset of experts to be fine-tuned based on their relevance scores. We define a threshold $p \in (0, 1]$ as a hyperparameter controlling the proportion of total relevance scores to be included in the selected subset. For each layer $l$, we select a set of top-scored experts $E_s^l$ whose cumulative relevance score exceeds the threshold $p$, satisfying:
\begin{equation}
\sum_{i \in E_s^l} R_{i}^l \geqslant p,
\end{equation}
where $R_{i}^l$ is the relevance score (either $r_i^l$ or $g_i^l$) of expert $i$ in layer $l$. During training and inference, tokens can be assigned to any expert. However, only the selected experts $E_s^l$ in each layer can be updated; other experts and modules remain frozen.


\section{Experiment Setup}
\subsection{Main Evaluation}
\label{sec:benchmarks}
We evaluate our ESFT method on two common LLM customization scenarios: (1) improving the model's \textbf{specific ability in a domain} where the model may already have decent performance; (2) adapting the model to a possibly \textbf{narrow but unfamiliar specialized task}.

\subsubsection{Tasks for Model Enhancement}
We choose two domain-specific tasks, i.e., Math and Code, to evaluate how our method can enhance the model's existing abilities. The two domains are widely concerned in current LLM research and suitable for evaluation, as many pre-trained models can perform decently, while there is significant potential for improvement through further training. We assess our method's effectiveness through performance gains.

For the Math domain, we use MetaMathQA~\cite{yu2023metamath} for training and use GSM8K~\cite{cobbe2021gsm8k} and MATH~\cite{hendrycks2021math} for evaluation. For the Code domain, We train the model on the Python subset of the enormous evol-codealpaca dataset~\cite{luo2023wizardcoder} to simulate a more concentrated LLM customization scenario, and assess its performance on HumanEval \cite{chen2021evaluating} and MBPP \cite{austin2021mbpp}. 

\begin{table*}[t]
    \centering
    \small
    \begin{tabular}{lccccccccc}
        \toprule
        & \multicolumn{2}{c}{\textbf{Math Ability}}
        & \multicolumn{2}{c}{\textbf{Code Ability}}
        & \multicolumn{4}{c}{\textbf{Specialized Tasks}} & \\
        \cmidrule(lr){2-3} \cmidrule(lr){4-5} \cmidrule(lr){6-9} 

        & MATH & GSM8K & Humaneval & MBPP & Intent & Summary & Law & Translation & Average \\
        \midrule
        Vanilla Model & 19.6 & 55.9 & \underline{42.1} & \underline{44.6} & 16.8 & 58.6  & 17.1 & 14.5 & 33.6 \\ \midrule
        FFT & \textbf{23.4} & \textbf{66.4} & \underline{42.1} & 42.2 & \textbf{78.8} & \textbf{69.4} & \underline{47.0} & \textbf{38.4} & \textbf{51.0} \\
        LoRA & 20.6 & 58.9 & 39.6 & \textbf{44.8} & 67.8 & 64.7 & 39.7 & 23.1 & 44.9 \\
        {ESFT-Token} (Ours) & 22.6 & \textbf{66.0} & 41.5 & 42.6 & 75.6 & 65.4 & 45.7 & \underline{36.2} & 49.4 \\
        {ESFT-Gate} (Ours) & \underline{23.2} & 64.9 & \textbf{43.3} & 41.8 & \textbf{78.6} & \underline{65.8} & \textbf{49.1} & 35.2 & \underline{50.2} \\
        \bottomrule
    \end{tabular}
    \caption{Main performance comparison across methods and tasks. Best or near-best results are shown in \textbf{bold} and second-best results are \underline{underlined}. Our method ESFT provides a strong balance of performance across diverse tasks, rivaling FFT and surpassing LoRA, particularly in specialized task domains.}
    \label{tab:performance}
\end{table*}

\subsubsection{Tasks for Model Adaptation}

We select four specialized tasks to evaluate how our method can facilitate language models to adapt to an unfamiliar downstream task, covering a diverse range of abilities that most models can excel at after training but not without training:
(1) Text-to-JSON Intent Recognition in the BDCI-21 Smart HCI NLU Challenge\footnote{\url{https://www.datafountain.cn/competitions/511}}, which requires converting text instructions into JSON format for home appliances. 
(2) Text Summarization in the BDCI-21 Summarization Challenge\footnote{\url{https://www.datafountain.cn/competitions/536}}, which summarizes customer service call transcripts.
(3) Legal judgment Prediction in the the BDCI-21 Law Event Prediction Challenge\footnote{\url{https://www.datafountain.cn/competitions/540}}, where the ``case description'' and ``judgment'' are repurposed as a legal judgment prediction task. (4) Low-resource Translation in the ChrEn dataset (\citealp{zhang2020chren}), translating the minority Cherokee to English. Examples of the tasks are shown in Appendix~\ref{app:examples}.

To measure model performance, for the text-to-JSON task, we calculate the exact match between model output and reference answer; for other tasks, we employ GPT-4 to score model output between 0 and 10 given reference answer\footnote{The exact version we use is \texttt{gpt-4-1106-preview}. The evaluation instructions are in Appendix~\ref{sec:app-evaluation}.}. All evaluations use few-shot examples.

\subsection{General Ability Evaluation}

We select a broad range of benchmarks to evaluate the extent to which the models' general abilities are preserved after training on new tasks.
These benchmarks include MMLU \cite{mmlu}, TriviaQA \cite{triviaqa}, HellaSwag \cite{hellaswag}, ARC-Challenge~\cite{arc}, IFEval~\cite{ifeval}, CEval~\cite{ceval}, and CLUEWSC~\cite{cluewsc}, covering comprehensive model ability evaluations across various domains including natural language understanding, question answering, instruction following, and common sense reasoning.

\begin{table*}[t]
    \centering
    \small
    \begin{tabular}{>{\arraybackslash}p{1.8cm} >{\centering\arraybackslash}p{1.35cm} >{\centering\arraybackslash}p{1.35cm} >{\centering\arraybackslash}p{1.35cm} >{\centering\arraybackslash}p{1.35cm} >{\centering\arraybackslash}p{1.35cm} >{\centering\arraybackslash}p{1.35cm} >{\centering\arraybackslash}p{1.35cm} >{\centering\arraybackslash}p{1.35cm}}
        \toprule
        & \textbf{CLUEWSC} & \textbf{TriviaQA} & \textbf{IFEval} & \textbf{MMLU} & \textbf{CEval} & \textbf{HellaSwag} & \textbf{ARC} & \textbf{Average} \\
        \midrule
        Vanilla Model & 81.5 & 67.7 & 42.5 & 57.5 & 59.9 & 74.0 & 53.7 & 62.4 \\ \midrule
        FFT & 80.9 $\pm$ 1.1 & 65.9 $\pm$ 0.7 & 34.2 $\pm$ 4.1 & 55.5 $\pm$ 1.0 & 58.8 $\pm$ 0.9 & 67.9 $\pm$ 3.8 & 48.4 $\pm$ 2.4 & 58.8 $\pm$ 1.3 \\
        LoRA & 74.3 $\pm$ 7.7 & 63.4 $\pm$ 5.4 & 38.7 $\pm$ 2.5 & 55.5 $\pm$ 1.2 & 57.0 $\pm$ 1.5 & \textbf{72.8} $\pm$ 1.9 & 51.8 $\pm$ 2.3 & 59.1 $\pm$ 2.5 \\
        ESFT-Token & 80.9 $\pm$ 0.9 & \textbf{66.7} $\pm$ 1.8 & \textbf{40.7} $\pm$ 1.3 & \textbf{57.1} $\pm$ 0.5 & \textbf{59.6} $\pm$ 0.8 & 72.3 $\pm$ 3.6 & \textbf{52.9} $\pm$ 1.5 & \textbf{61.5} $\pm$ 1.1 \\
        ESFT-Gate & \textbf{81.4} $\pm$ 1.1 & \textbf{66.5} $\pm$ 2.3 & 40.2 $\pm$ 1.5 & \textbf{57.0} $\pm$ 0.4 & \textbf{59.5} $\pm$ 0.8 & 68.2 $\pm$ 9.9 & 51.5 $\pm$ 3.1 & 60.6 $\pm$ 2.3 \\
        \bottomrule
    \end{tabular}
    \caption{General ability performance comparison across methods and tasks. The performance for a task is averaged across all training experiments, followed by the standard deviation across tasks. Best or near-best results are shown in \textbf{bold}. Our method ESFT consistently achieves good performance among all tasks. }
    \label{tab:general}
\end{table*}

\subsection{Backbone Model and Training Settings}
We use the backbone architecture of DeepSeek-V2-Lite \cite{Model_DeepSeekV2} for all experiments. 
The model includes a fine-grained set of 66 experts for each transformer layer. This makes it uniquely suitable at the time of this study for our method, which benefits from expert specialization. 
We train the model on a carefully curated alignment dataset that excludes math and code data and take the resulting checkpoint as our vanilla model for subsequent experiments. This alignment phase can activate model ability across various domains while keeping Math/Code ability as elementary to better verify the performance gains of our method in these two fields.

We adopt two baselines: Full-Parameter Fine-Tuning (FFT) and Low-Rank Adaptation (LoRA, \citealp{5-LoRA}). For LoRA, we add low-rank matrices to all parameters for training except token embeddings and the language modeling head. 
We maintain a 1:1 ratio for task-specific data and alignment data for all methods, which we find is highly effective in preserving general abilities obtained from the alignment phase for FFT and LoRA. 
However, for our ESFT method, not adopting this data mixing strategy may even better maintain general ability. We detail this in Appendix~\ref{app:sft-mix}. 
All experiments are done on the HFAI cluster\footnote{\href{https://doc.hfai.high-flyer.cn/index.html}{https://doc.hfai.high-flyer.cn/index.html}} with 2 nodes of 8x Nvidia A100 PCIe GPUs. 

For hyperparameter settings, all methods use a batch size of 32 and a sequence length of 4096 for training. For every task, we set the maximum steps of training to 500, and evaluate the model every 100 steps. The learning rates are set to 3e-5, 1e-4, and 1e-5 for FFT, LoRA, and ESFT, respectively, based on a hyperparameter search in \{1e-5, 3e-5, 1e-4, 3e-4\}. The LoRA rank is set to 8 and scaling is set to 2, following~\citet{5-LoRA}. The threshold $p$ is set to 0.1 for ESFT-Gate and 0.2 for ESFT-Token, respectively. \S\ref{sec:param} shows how we determine the threshold for ESFT.

\section{Results}

\subsection{Benchmark Performance Results}
\label{sec:mainresult}
The results in Table~\ref{tab:performance} and Table~\ref{tab:general} demonstrate several conclusions. All methods can improve model performance in customization tasks compared to the vanilla model, while they may cause a performance decrease in general tasks. Generally, the performance increase is higher in model adaptation tasks than in model enhancement tasks.

For customization ability evaluation, ESFT surpasses LoRA significantly and is competitive with FFT. As shown in Table~\ref{tab:performance}, ESFT-Token and ESFT-Gate achieve near-best results in model enhancement tasks like Math, and ESFT-Gate achieves the best performance in the Humaneval task. ESFT also excels in model adaptation tasks, with ESFT-Gate achieving near-best performance in 3 tasks out of 4. Notably, ESFT-Gate's average of 50.2 is competitive compared to FFT's 51.0, slightly better than ESFT-Token's 49.4, and significantly surpasses LoRA's 44.9. This demonstrates that finding task-relevant experts can efficiently adapt the model for efficient customization.

For general ability evaluation, ESFT consistently outperforms FFT and LoRA by showing less performance degradation. As illustrated in Table \ref{tab:general}, ESFT-token performs better than ESFT-gate, with average scores of 61.5 and 60.6, respectively. The results demonstrate a wide range of retention in tasks such as TriviaQA and IFEval, surpassing FFT's 58.8 and LoRA's 59.1. Both methods retain performance better than LoRA and FFT, highlighting their effectiveness in maintaining general task performance\footnote{
We further investigate Math and Code performance of the models trained on specialized tasks in Appendix~\ref{app:mathcode}. FFT and LoRA exhibit even more severe degradation, while ESFT shows a minimal performance drop. 
}. Analyses in \S\ref{sec:abl-shared} indicate that 
such degradation on general tasks for FFT and LoRA may result from training shared parameters.

\subsection{Computational Efficiency Results}
\label{sec:efficiencyresults}
The results in Figure~\ref{fig:eff} demonstrates that ESFT exhibits several advantages in terms of training time and storage space requirements:

\begin{figure*}[h]
    \centering
    \includegraphics[width=1.07\textwidth, trim=50 0 0 0, clip]{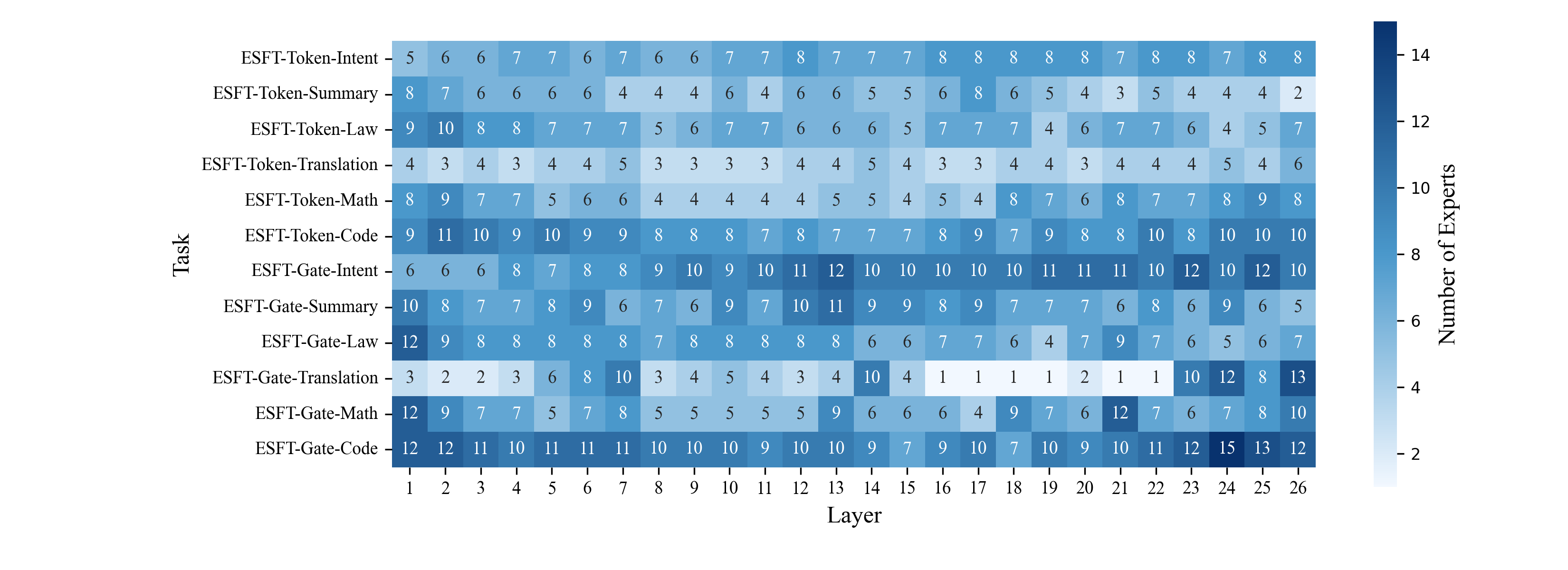}
    \caption{Number of experts trained in ESFT across layers and tasks. Earlier computed layers are numbered smaller. Most tasks and layers train 5-15\% of experts, demonstrating ESFT's effectiveness in selecting task-related experts.}
    \label{fig:esft_experts_heatmap}
\end{figure*}

\begin{figure}[t]
    \centering
    \includegraphics[width=\linewidth]{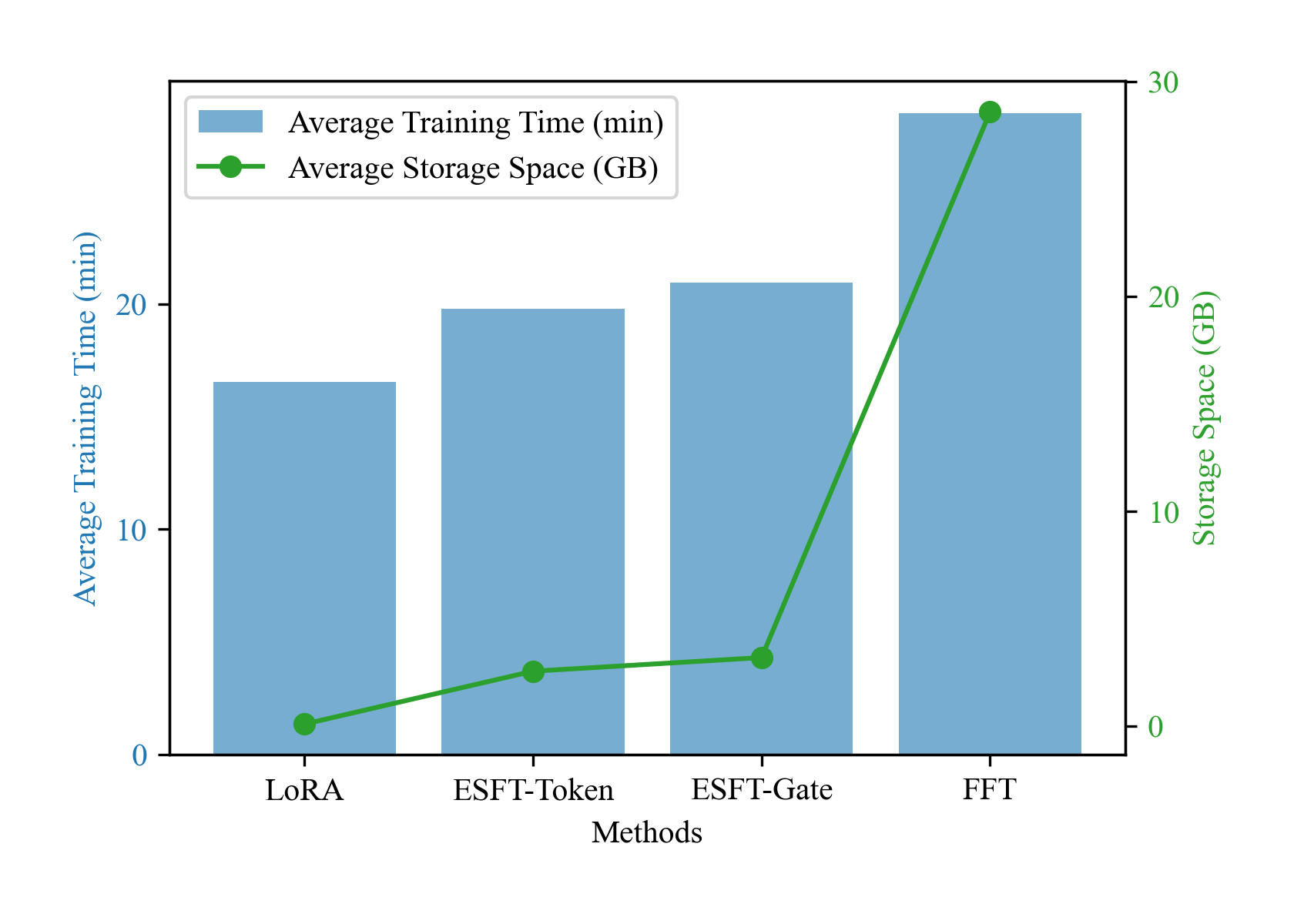}
    \caption{Computational efficiency results. Blue bars show the training time and green lines show storage space. ESFT performs efficiently in terms of training time and storage space.}
    \label{fig:eff}
\end{figure}

\paragraph{Training Time}
The average training time for ESFT-Token and ESFT-Gate is 19.8 minutes and 20.9 minutes, respectively. The FFT method takes significantly longer at 28.5 minutes. Although LoRA achieves a shorter training time of 16.5 minutes, our methods are relatively close.

\paragraph{Storage Space}
The average storage space of parameters trained is 2.57 GB for ESFT-Token and 3.20 GB for ESFT-Gate, while FFT demands a substantial 28.6 GB.
Although LoRA requires less storage,
ESFT performs significantly better than LoRA in downstream task performance.


In summary, ESFT demonstrates excellent performance in training time and storage space, significantly outperforming FFT. 
Furthermore, as shown in Table~\ref{tab:ablation-shared}, ESFT requires much fewer trainable parameters compared to FFT, resulting in lower GPU memory usage.
These advantages show that ESFT is efficient and effective for language model customization and adaptation.

\section{Analysis}

\begin{table*}[t]
    \centering
    \small
    \begin{tabular}{>{\centering\arraybackslash}p{1.6cm} >{\centering\arraybackslash}p{1.6cm} >{\centering\arraybackslash}p{1.62cm} >{\centering\arraybackslash}p{1.62cm} >{\centering\arraybackslash}p{1.62cm} >{\centering\arraybackslash}p{1.62cm} >{\centering\arraybackslash}p{1.62cm}}
        \toprule
        \textbf{Non-shared Experts} & \textbf{Shared Experts} & \textbf{Non-expert Parameters} & \textbf{Trainable Parameters} & \textbf{Specialized Ability} & \textbf{General Ability} & \textbf{Average}\\
        \midrule
         ALL & $\checkmark$ & $\checkmark$ & 15.7B  & \textbf{51.0} & 58.8 & 54.9\\
        Relevant & $\checkmark$ & $\times$ & 1.85B & 49.8 & 60.7 & 55.3\\
        Relevant & $\times$ & $\times$ & 1.4B & 49.4 & \textbf{61.5} & \textbf{55.4}\\
        $\times$ & $\checkmark$ & $\times$ & \textbf{450M} & 47.4 & 61.2 & 54.3\\
        $\times$ & $\checkmark$ & $\checkmark$ & 1.3B & 49.0 & 60.0 & 54.5\\
        Relevant & $\checkmark$ & $\checkmark$ & 2.7B & \textbf{50.8} & 60.3 & \textbf{55.6}\\
        \midrule
         $\times$ & $\times$ & $\times$ & - & 33.8 & 62.4 & 48.1\\
        \bottomrule
    \end{tabular}
    \caption{Comparisons of different model configs based on whether training shared or non-shared parameters. Results include trainable parameters and performance of specialized and general abilities. The best or near-best results excluding the non-training setting are shown in \textbf{bold}.}
    \label{tab:ablation-shared}
\end{table*}
\begin{figure*}[t]
        \centering
        \includegraphics[width=\linewidth]{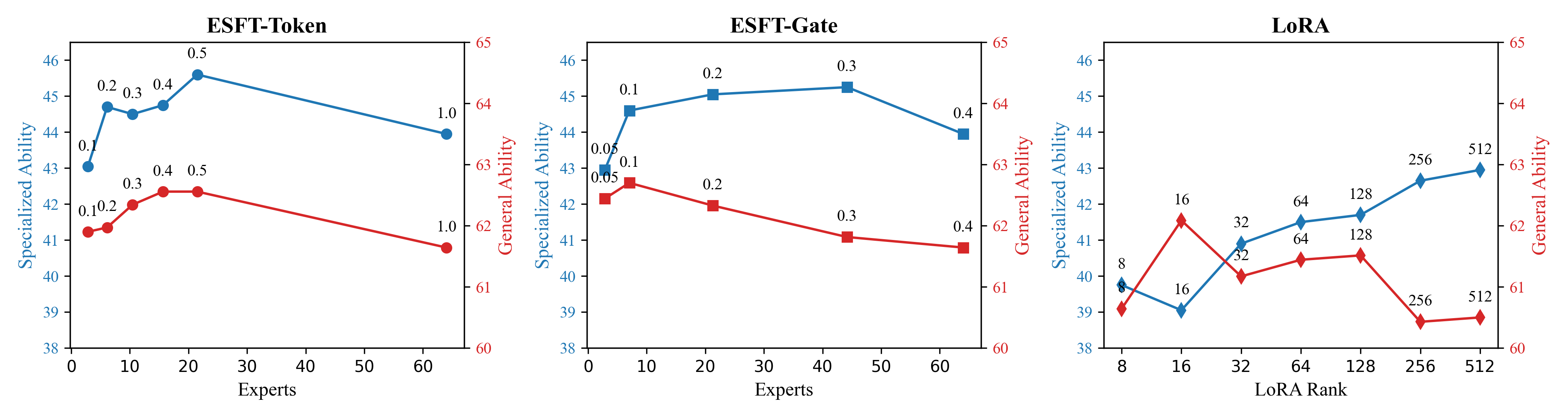}
        \caption{Comparison of three methods under different training efficiency settings on the Math task. The x-axis shows the average trainable experts per layer for ESFT and rank for LoRA, indicating the ratio of trained parameters. The y-axis represents specialized and general ability. Markers on the lines indicate $p$ or rank values. ESFT consistently outperforms LoRA in both specialized and general ability. }
        \label{fig:eff}
        
\end{figure*}

 
In this section, we investigate the expert selection process of ESFT in \S\ref{sec:epselection}, and demonstrate the performance of ESFT and LoRA under different computational constraints in \S\ref{sec:param}. We analyze the effects of training shared and non-shared parameters in \S\ref{sec:abl-shared}, and conduct ablation studies in \S\ref{sec:abl} to verify the importance of our expert relevance scores and model structure of fine-grained experts.

\subsection{ESFT Leverages Specialized Experts Effectively}
\label{sec:epselection}

We analyze the number of experts ESFT trains across tasks and layers to understand its expert selection process. Results are shown in Figure \ref{fig:esft_experts_heatmap}.

From the results, we have several observations: (1) The average number of experts used per task across layers ranges from 2 to 15 out of 66, indicating ESFT can have 75\%-95\% fewer trainable parameters than FFT. (2) ESFT-Token generally employs fewer experts while better maintaining general performance, comparable to ESFT-Gate in tasks like Math, Intent, and Law. (3) The number of experts varies by task, with more specialized tasks like Math and Translation using fewer experts; our method's performances for these tasks exceed LoRA to the greatest extent, indicating that our method is especially suitable for more specialized tasks. (4) For most tasks, few experts are chosen in the middle layers, indicating that expert distribution is more concentrated in these layers.

\subsection{ESFT Leverages Training Resources Efficiently}

\label{sec:param}

\begin{table*}[t]
    \centering
    \small
    \begin{tabular}{lccccccccc}
        \toprule
        & \multicolumn{2}{c}{\textbf{Math Ability}}
        & \multicolumn{2}{c}{\textbf{Code Ability}}
        & \multicolumn{4}{c}{\textbf{Specialized Tasks}} & \\
        \cmidrule(lr){2-3} \cmidrule(lr){4-5} \cmidrule(lr){6-9} 
        & MATH & GSM8K & Humaneval & MBPP & Intent & Summary & Law & Translation & Average \\
        \midrule
        {ESFT-Token} & 22.6 & 66.0 & 41.5 & 42.6 & 75.6 & 65.4 & 45.7 & 36.2 & 49.4 \\
        ~~$\Delta$ \texttt{of rand} & -1.0 & -3.7 & -2.5 & 0.2 & -2.6 & -1.7 & 1.3 & -13.5 & -2.8  \\        
        ESFT-Gate & 23.2 & 64.9 & 43.3 & 41.8 & 78.6 & 65.8 & 49.1 & 35.2 & 50.2 \\
        ~~$\Delta$ \texttt{of rand} & -1.7 & -3.2 & -4.3 & 1.6 & -5.0 & 0.3 & -2.9 & -20.4 & -4.4  \\
        \bottomrule
    \end{tabular}
    \caption{Performance comparison between original experts and random experts. Replacing high-affinity experts with random ones significantly harms model performance across different tasks.}
    \label{tab:random_experts}
\end{table*}

Both ESFT and LoRA have a training efficiency hyperparameter ($p$ for ESFT and rank for LoRA). Increasing its value would raise computational resource usage and potentially improve performance. To understand how ESFT and LoRA perform under different efficiency settings, we evaluate benchmark performance on the Math task. We set rank $\leqslant$ 512 for LoRA as a higher value will result in more trainable parameters than FFT.
Figure~\ref{fig:eff} illustrates both specialized and general ability under different training efficiency settings. 

From the results, we can conclude: (1) All three methods show a trade-off between training efficiency and performance. Increasing trained parameters ($p$ for ESFT and rank for LoRA) before a certain point can improve performance. (2) Both ESFT-Token and ESFT-Gate outperform LoRA at any point, demonstrating higher specialized ability and more stable general ability. (3) ESFT-Token peaks in both specialized and general ability at $p$=0.5, while ESFT-Gate peaks at $p$=0.3 for specialized and $p$=0.1 for general ability. (4) ESFT-Token and ESFT-Gate performance saturates at $p$=0.2 and $p$=0.1, respectively, indicating that most expert choices may be less relevant to task performance. We delve deeper into this in Appendix~\ref{app:qualitative}. 


\subsection{Selectively Training Non-Shared Parameters is the Key to ESFT}
\label{sec:abl-shared}

In our proposed ESFT method, we only fine-tune a subset of non-shared experts. This section provides detailed discussions of several variants of our method that may also train \textit{shared} parameters. The variables are based on:
\begin{itemize}
\item Whether \textbf{all} non-shared experts or a \textbf{task-relevant} subset of them (we use the Token Selection Ratio and set $p$=0.2) are trained.
\item Whether shared experts are trained.
\item Whether other parameters, including gates, attention layers, and embeddings, are trained.
\end{itemize}

The results are shown in Table~\ref{tab:ablation-shared}. We report average trainable parameters across all tasks, performance of specialized and general abilities, and their average. Detailed numbers for all benchmarks are shown in Appendix~\ref{app:ablation-shared}.
From the results, we can draw several conclusions:

\textbf{Specialized performance increases as trainable parameters increase.} The rank of trainable parameters from 450M to 15.7B highly aligns with the rank of specialized ability from 47.4 to 51.0. This suggests that increasing trainable parameters is effective in enhancing specialized performance.

\textbf{General performance decreases as trainable \textit{shared} parameters increase.} Whether relevant non-shared experts are trained or not, general performance decreases from 61.5 to 60.3, or from 62.4 to 60.0, respectively, as we train shared experts and/or non-expert parameters. As the complete set of non-shared experts is trained, general performance decreases further from 60.3 to 58.8. This suggests that training shared parameters is more likely to cause overfitting on downstream tasks and forgetting on general tasks compared to training non-shared parameters.

\textbf{It is highly prioritized to train task-relevant non-shared experts.} Training relevant experts achieves at least 55.3, while other settings achieve at most 54.9, even with higher demands of up to 15.7B parameters. Therefore, fine-tuning these experts is highly prioritized for model customization.

We propose two major training strategies based on these conclusions:
\begin{enumerate}
\item \textbf{Prioritize specialized ability:} Train all shared parameters and task-relevant non-shared experts to maximize the enhancement of specialized performance.
\item \textbf{Balance specialized and general ability, and computational efficiency:} Train only task-relevant non-shared experts to minimize parameter costs while maximizing the maintenance of general ability.
\end{enumerate}

\begin{figure}[t]
        \centering
        \includegraphics[width=\linewidth]{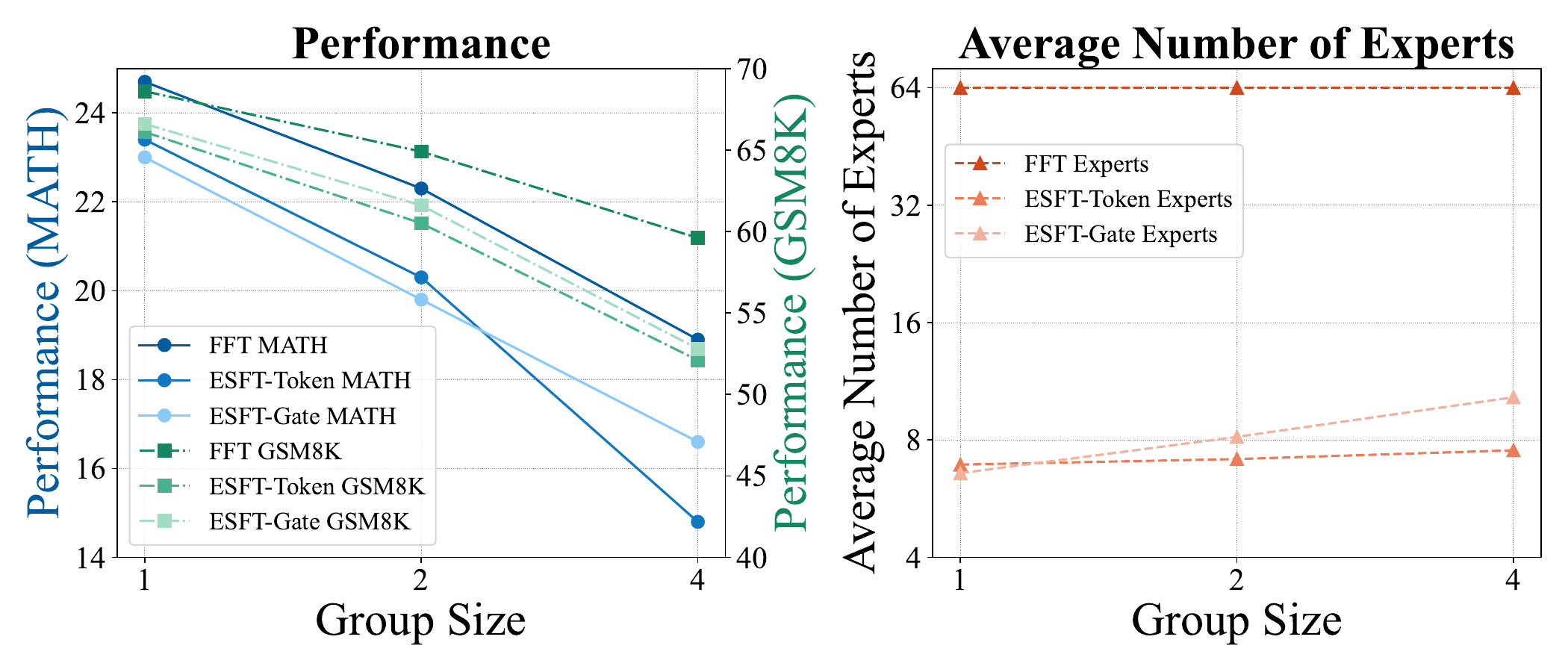}
        \caption{Experiment results for grouped experts. As the experts become more coarse-grained, ESFT degrades more severely than FFT.}
        \label{fig:bindexperts}
        
\end{figure}

\subsection{Analysis of Key Modules in ESFT}
In this section, we analyze and demonstrate that the effectiveness of our method lies in two modules: (1) our proposed expert relevance score functions and (2) the fine-grained expert segmentation of the MoE model architecture.

\label{sec:abl}

\paragraph{Expert Relevance Score Function} In this work, we propose Average Gate Score and Token Selection Ratio as expert relevance score functions to filter relevant experts for different tasks. To demonstrate their effectiveness, we replace the experts obtained from these functions with random experts while keeping the number of activated experts per layer the same. Results in Table~\ref{tab:random_experts} show that replacing relevant experts with random ones significantly decreases task performance, demonstrating the effectiveness of our proposed relevance scores.

\paragraph{Fine-Grained Expert Segmentation of the MoE Model}
We use the fine-grained segmented DeepSeek-V2 model as our backbone. To demonstrate t the effectiveness of this fine-grained segmentation, we use greedy search (as detailed in Appendix~\ref{app:greedy}) to group experts, simulating coarse-grained segmentation. Experts in the same group share the average affinity score. We maintain the computational cost by selecting a constant 1/8 of experts for each token. Experiment results of the Math domain in Figure~\ref{fig:bindexperts} show that as the group size increases, our method's performance decreases more severely than FFT, while the training cost (i.e., trainable experts) rises. These findings indicate that our method, and even effective LLM customization, highly rely on a fine-grained segmented LLM architecture with more specialized experts.

\section{Conclusion}
In this work, we study parameter-efficient fine-tuning methods for sparse large language models with the Mixture of Experts (MoE) architecture. We first observe that tasks from different domains are handled by distinct combinations of experts. We then propose selecting the most relevant experts for downstream tasks using two metrics: average gate score and token selection ratio. Experimental results show that our method significantly reduces training costs while matching or surpassing full parameter fine-tuning results. Further analysis confirms that our method enhances the specialization of the expert system within the MoE architecture.

\section*{Limitations}
Firstly, due to the limitation of the availability of other fine-grained MoE models, our method was only tested on the DeepSeek-V2-Lite MoE model. The conclusions drawn from this model require further validation when applied to other contexts.
Besides, due to the lack of parameter-wise and structurally aligned MoE models with different expert granularities, we used a simulation approach by binding several groups of experts to compare coarse-grained and fine-grained MoE methods.

\bibliography{custom}
\newpage
~
\newpage
\appendix

\begin{table*}[t]
    \centering
    \includegraphics[width=\linewidth]{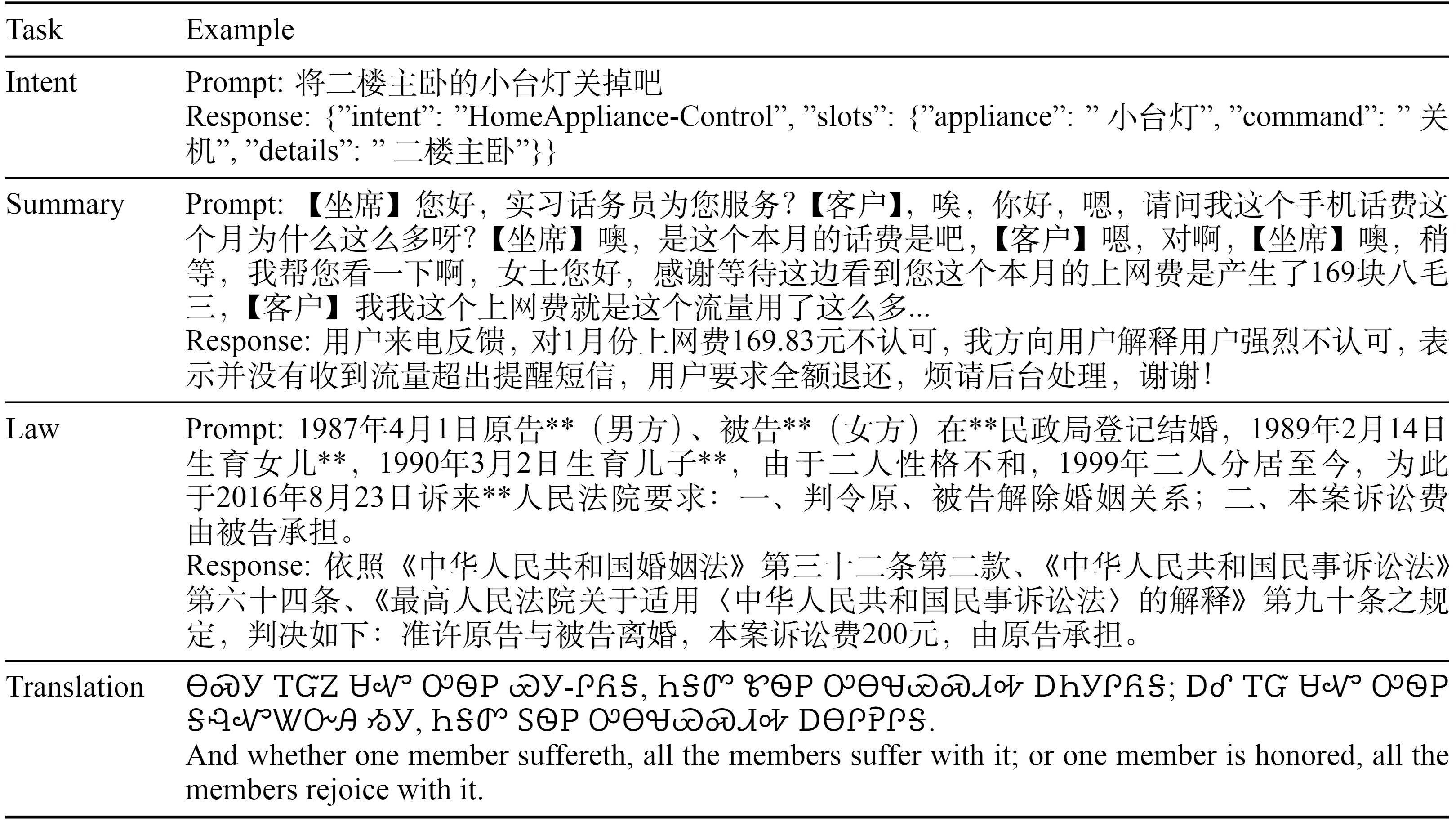}
    \caption{Examples for different specialized tasks.}
    \label{tab:task_descriptions}
    \end{table*}


\begin{table*}[t]
    \centering
    \scriptsize
    \begin{tabular}{>{\centering\arraybackslash}p{0.7cm} >{\centering\arraybackslash}p{0.7cm} >{\centering\arraybackslash}p{0.7cm} >{\centering\arraybackslash}p{1.1cm} >{\centering\arraybackslash}p{1.1cm} >{\centering\arraybackslash}p{1.1cm} >{\centering\arraybackslash}p{1.1cm} >{\centering\arraybackslash}p{1.1cm} >{\centering\arraybackslash}p{1.3cm} >{\centering\arraybackslash}p{1.1cm} >{\centering\arraybackslash}p{1.1cm}}
        \toprule
        \textbf{Non-shared} & \textbf{Shared} & \textbf{Non-expert} & \textbf{CLUEWSC} & \textbf{TriviaQA} & \textbf{IFEval} & \textbf{MMLU} & \textbf{CEval} & \textbf{HellaSwag} & \textbf{ARC} & \textbf{Average} \\\midrule
        ALL & $\checkmark$ & $\checkmark$ & \textbf{80.9} $\pm$ 2.2 & 65.9 $\pm$ 1.5 & 34.2 $\pm$ 8.1 & 55.5 $\pm$ 1.9 & 58.8 $\pm$ 1.7 & 67.9 $\pm$ 7.4 & 48.4 $\pm$ 4.7 & 58.8 $\pm$ 2.5 \\ 
        Relevant & $\checkmark$ & $\times$ & \textbf{80.9} $\pm$ 2.1 & 66.1 $\pm$ 4.4 & \textbf{42.4} $\pm$ 3.0 & 56.8 $\pm$ 1.0 & 58.9 $\pm$ 1.6 & 67.8 $\pm$ 20.4 & 52.1 $\pm$ 5.7 & 60.7 $\pm$ 4.4 \\
        Relevant & $\times$ & $\times$ & \textbf{80.9} $\pm$ 1.8 & \textbf{66.7} $\pm$ 3.5 & 40.7 $\pm$ 2.6 & \textbf{57.1} $\pm$ 1.0 & \textbf{59.6} $\pm$ 1.5 & \textbf{72.3} $\pm$ 7.0 & \textbf{52.9} $\pm$ 3.0 & \textbf{61.5} $\pm$ 2.3 \\
        $\times$ & $\checkmark$ & $\times$ & \textbf{81.1} $\pm$ 3.4 & \textbf{66.7} $\pm$ 4.2 & 41.2 $\pm$ 1.6 & 56.9 $\pm$ 1.2 & 58.9 $\pm$ 1.6 & 71.3 $\pm$ 14.1 & 52.6 $\pm$ 5.6 & 61.2 $\pm$ 3.3 \\
        $\times$ & $\checkmark$ & $\checkmark$ & 79.5 $\pm$ 4.4 & 65.8 $\pm$ 5.0 & 41.4 $\pm$ 3.2 & 56.2 $\pm$ 1.6 & 58.6 $\pm$ 1.7 & 67.5 $\pm$ 20.7 & 51.2 $\pm$ 4.1 & 60.0 $\pm$ 4.4 \\ 
        Relevant & $\checkmark$ & $\checkmark$ & 80.4 $\pm$ 4.1 & 66.3 $\pm$ 4.1 & 41.1 $\pm$ 5.0 & 56.7 $\pm$ 1.2 & 59.0 $\pm$ 1.9 & 67.5 $\pm$ 20.3 & 51.5 $\pm$ 4.6 & 60.3 $\pm$ 4.6 \\\midrule
        $\times$ & $\times$ & $\times$ & 81.5 & 67.7 & 42.5 & 57.5 & 59.9 & 74.0 & 53.7 & 62.4 \\
        \bottomrule
    \end{tabular}
    \caption{Performance of general tasks across methods based on whether training shared or non-shared parameters. The performance for a task is averaged across all training experiments, followed by the standard deviation across tasks. Best or near-best results are shown in \textbf{bold}. }
    \label{tab:general_sharednonshared}
\end{table*}

\begin{table*}[t]
    \centering
    \scriptsize
    \begin{tabular}{>{\centering\arraybackslash}p{0.9cm} >{\centering\arraybackslash}p{0.9cm} >{\centering\arraybackslash}p{0.9cm} >{\centering\arraybackslash}p{0.9cm} >{\centering\arraybackslash}p{0.9cm} >{\centering\arraybackslash}p{0.9cm} >{\centering\arraybackslash}p{0.9cm} >{\centering\arraybackslash}p{0.8cm} >{\centering\arraybackslash}p{0.8cm} >{\centering\arraybackslash}p{0.8cm} >{\centering\arraybackslash}p{0.9cm} >{\centering\arraybackslash}p{0.9cm}}
        \toprule
        \multirow{2}{*}{\centering\arraybackslash\textbf{Non-shared}} & \multirow{2}{*}{\centering\arraybackslash\textbf{Shared}} & \multirow{2}{*}{\centering\arraybackslash\textbf{Non-expert}}& \multicolumn{2}{c}{\textbf{Math Ability}}
        & \multicolumn{2}{c}{\textbf{Code Ability}}
        & \multicolumn{4}{c}{\textbf{Specialized Tasks}} & \\
        \cmidrule(lr){4-5} \cmidrule(lr){6-7} \cmidrule(lr){8-11} 
        &&& MATH & GSM8K & Humaneval & MBPP & Intent & Summary & Law & Translation & Average \\        
        \midrule
        ALL & $\checkmark$ & $\checkmark$ & 23.4 & \textbf{66.4} & 42.1 & 42.2 & 78.8 & \textbf{69.4} & \textbf{47.0} & \textbf{38.4} & \textbf{51.0} \\
        Relevant & $\checkmark$ & $\times$ & 23.8 & 65.7 & 40.2 & 43.8 & 80.4 & 67.3 & 42.4 & 35.1 & 49.8 \\
        Relevant & $\times$ & $\times$ & 22.6 & 66.0 & 41.5 & 42.6 & 75.6 & 65.4 & 45.7 & 36.2 & 49.4 \\
        $\times$ & $\checkmark$ & $\times$ & 22.7 & 64.5 & 37.2 & 44.0 & 73.6 & 68.3 & 42.7 & 26.0 & 47.4 \\
        $\times$ & $\checkmark$ & $\checkmark$ & 23.4 & \textbf{66.6} & 41.5 & 44.4 & 81.0 & 66.7 & 39.0 & 29.5 & 49.0 \\
        Relevant & $\checkmark$ & $\checkmark$ & \textbf{24.8} & 66.0 & 42.1 & 43.2 & \textbf{82.2} & \textbf{69.5} & 46.4 & 32.2 & \textbf{50.8} \\
        \midrule
        $\times$ & $\times$ & $\times$ & 19.6 & 55.9 & 42.1 & \textbf{44.6} & 16.8 & 58.6 & 17.1 & 14.5 & 33.6 \\
        \bottomrule
    \end{tabular}
    \caption{Performance of specialized tasks across methods based on whether training shared or non-shared parameters. Best or near-best results are shown in \textbf{bold}.}
    \label{tab:specialized_sharednonshared}
\end{table*}

\begin{table*}[h]
    \centering
    \scriptsize
    \begin{tabular}{lccccc}
        \toprule
        & \multicolumn{2}{c}{\textbf{Math Ability}}
        & \multicolumn{2}{c}{\textbf{Code Ability}} & \\
        \cmidrule(lr){2-3} \cmidrule(lr){4-5} 

        & MATH & GSM8K & HumanEval & MBPP & Average \\
        \midrule
        Vanilla Model & 19.6 & 55.9 & 42.1 & 44.6 & 40.5 \\
        \midrule
        FFT & 15.1 $\pm$ 0.3 & 40.3 $\pm$ 5.3 & 30.2 $\pm$ 4.4 & 40.6 $\pm$ 3.9 & 31.5 $\pm$ 2.5 \\
        LoRA & 11.8 $\pm$ 0.6 & 36.1 $\pm$ 4.4 & 27.9 $\pm$ 2.3 & 36.6 $\pm$ 2.6 & 28.1 $\pm$ 2.0 \\
        ESFT-Token & \textbf{19.4} $\pm$ 0.8 & \textbf{55.2} $\pm$ 0.7 & \textbf{39.5} $\pm$ 1.0 & 44.8 $\pm$ 0.8 & \textbf{39.7} $\pm$ 0.4 \\
        ESFT-Gate & \textbf{19.5} $\pm$ 0.3 & \textbf{55.1} $\pm$ 1.3 & \textbf{39.3} $\pm$ 1.3 & \textbf{45.3} $\pm$ 0.6 & \textbf{39.8} $\pm$ 0.6 \\
        \bottomrule
    \end{tabular}
    \caption{Math and Code performance comparison across methods trained on specialized tasks. Best or near-best results are shown in \textbf{bold}. ESFT retains performance significantly better  compared to FFT and LoRA.}
    \label{tab:mathcode}
\end{table*}

\begin{table*}[t]
    \centering
    \small
    \begin{tabular}{lccccccccc}
        \toprule
        & \multicolumn{2}{c}{\textbf{Math Ability}}
        & \multicolumn{2}{c}{\textbf{Code Ability}}
        & \multicolumn{4}{c}{\textbf{Specialized Tasks}} & \\
        \cmidrule(lr){2-3} \cmidrule(lr){4-5} \cmidrule(lr){6-9} 

        & MATH & GSM8K & HumanEval & MBPP & Intent & Service & Law & Translation & Average \\
        \midrule
        FFT & 26.1 & 70.4 & 51.2 & 42.6 & 78.8 & 72.8 & 45.6 & 34.4 & 52.7 \\
        \texttt{~+ mix data} & -2.7 & -4.0 & -9.1 & -0.4 & 0.0 & -3.4 & 1.4 & 4.0 & -1.7 \\
        LoRA & 21.8 & 57.8 & 42.1 & 42.6 & 78.2 & 66.4 & 46.0 & 21.8 & 47.1 \\
        \texttt{~+ mix data} & -1.2 & 1.1 & -2.5 & 2.2 & -10.4 & -1.7 & -6.3 & 1.3 & -2.2 \\
        ESFT-Token & 25.2 & 64.8 & 42.1 & 43.8 & 78.0 & 67.4 & 47.2 & 31.9 & 50.0 \\
        \texttt{~+ mix data} & -2.6 & 1.2 & -0.6 & -1.2 & -2.4 & -2.0 & -1.5 & 4.3 & -0.6 \\
        ESFT-Gate & 24.1 & 64.9 & 42.1 & 44.6 & 77.2 & 68.4 & 43.6 & 32.8 & 49.7 \\
        \texttt{~+ mix data} & -0.9 & 0.0 & 0.0 & -2.8 & 1.4 & -2.6 & 0.9 & 2.4 & 0.5 \\
        \bottomrule
    \end{tabular}
    \caption{Downstream task performance comparison across methods and tasks with and without mixing data from the alignment phase. Results show that mixing alignment data leads to a minor performance decrease for most methods.}
    \label{tab:mix-main}
\end{table*}

\begin{table*}[t]
    \centering
    \small
    \begin{tabular}{>{\arraybackslash}p{1.8cm} >{\centering\arraybackslash}p{1.35cm} >{\centering\arraybackslash}p{1.35cm} >{\centering\arraybackslash}p{1.35cm} >{\centering\arraybackslash}p{1.35cm} >{\centering\arraybackslash}p{1.35cm} >{\centering\arraybackslash}p{1.35cm} >{\centering\arraybackslash}p{1.35cm} >{\centering\arraybackslash}p{1.35cm}}
        \toprule
        & \textbf{CLUEWSC} & \textbf{TriviaQA} & \textbf{IFEval} & \textbf{MMLU} & \textbf{CEval} & \textbf{HellaSwag} & \textbf{ARC} & \textbf{Average} \\
        \midrule
        Vanilla Model & 81.5 & 67.7 & 42.5 & 57.5 & 59.9 & 74.0 & 53.7 & 62.4 \\ \midrule
        FFT & 76.8 $\pm$ 1.7 & 62.4 $\pm$ 10 & 28.4 $\pm$ 5.1 & 55.5 $\pm$ 1.1 & 58.4 $\pm$ 0.4 & 74.6 $\pm$ 3.2 & 53.6 $\pm$ 3.1 & 58.5 $\pm$ 2.5 \\
        \texttt{~+ mix data} & 4.1 & 3.5 & 5.8 & 0.0 & 0.4 & -6.7 & -5.2 & 0.3 \\
        LoRA & 60.2 $\pm$ 27 & 61.2 $\pm$ 4.0 & 33.4 $\pm$ 6.1 & 52.3 $\pm$ 3.3 & 55.3 $\pm$ 2.3 & 71.5 $\pm$ 2.5 & 50.7 $\pm$ 2.2 & 55.0 $\pm$ 4.6 \\
        \texttt{~+ mix data} & 14.1 & 2.2 & 5.3 & 3.2 & 1.7 & 1.3 & 1.1 & 4.1 \\
        ESFT-Token & 80.0 $\pm$ 2.5 & 67.5 $\pm$ 0.3 & 41.9 $\pm$ 0.8 & 57.3 $\pm$ 0.2 & 60.2 $\pm$ 0.5 & 74.5 $\pm$ 0.7 & 54.9 $\pm$ 0.7 & 62.3 $\pm$ 0.5 \\
        \texttt{~+ mix data} & 0.9 & -0.8 & -1.2 & -0.2 & -0.6 & -2.2 & -2.0 & -0.8 \\
        ESFT-Gate & 80.2 $\pm$ 1.6 & 67.6 $\pm$ 0.3 & 40.8 $\pm$ 2.4 & 57.3 $\pm$ 0.3 & 59.9 $\pm$ 0.4 & 74.3 $\pm$ 0.9 & 55.1 $\pm$ 0.9 & 62.2 $\pm$ 0.5 \\
        \texttt{~+ mix data} & 1.2 & -1.1 & -0.6 & -0.3 & -0.4 & -6.1 & -3.6 & -1.6 \\
        \bottomrule
    \end{tabular}
    \caption{General task performance comparison across methods and tasks with and without alignment data mixing. Results show that mixing alignment data improves FFT and LoRA in general tasks, but not our ESFT method. It showcases that ESFT can adapt to downstream tasks directly with minimal performance loss in general tasks.}
    \label{tab:mix-general}
\end{table*}

\begin{CJK}{UTF8}{gbsn}
\begin{table*}[t]
\centering
\small
\begin{tabular}{@{}lp{13cm}@{}}
\toprule
Task     & Evaluation Instruction \\ \midrule
Summary       & 请你进行以下电话总结内容的评分。请依据以下标准综合考量，以确定预测答案与标准答案之间的一致性程度。满分为10分，根据预测答案的准确性、完整性和相关性来逐项扣分。请先给每一项打分并给出总分，再给出打分理由。总分为10分减去每一项扣除分数之和，最低可扣到0分。请以“内容准确性扣x分，详细程度/完整性扣x分，...，总分是：x分"为开头。 1. \textbf{内容准确性}： - 预测答案是否准确反映了客户问题或投诉的核心要点。 - 是否有任何关键信息被错误陈述或误解。 2. \textbf{详细程度/完整性}： - 预测答案中包含的细节是否充分，能否覆盖标准答案中所有重要点。 - 对于任何遗漏的关键信息，应相应减分。 3. \textbf{内容冗余度}： - 预测答案是否简洁明了，和标准答案风格一致，不存在冗余信息。 - 如果预测答案过长或与标准答案风格不一致，需相应减分。 4. \textbf{行动指令正确性}： - 预测答案对后续处理的建议或请求是否与标准答案相符。 - 如果处理建议发生改变或丢失，需相应减分。 预测答案：\{prediction\} 参考答案：\{ground\_truth\} \\ \midrule
Law & 请你进行以下法案判决预测内容的评分。请依据以下标准综合考量，以确定预测答案与标准答案之间的一致性程度。满分为10分，根据预测答案的准确性、完整性和相关性来逐项扣分。请先给每一项打分并给出总分，再给出打分理由。总分为10分减去每一项扣除分数之和，最低可扣到0分。请以“相关性扣x分，完整性扣x分，...，总分是：x分"为开头。 1. \textbf{相关性}：预测答案与标准答案的相关程度是最重要的评分标准。如果预测的判决情况与标准答案完全一致，即所有事实和结果都被精确复制或以不同但等效的方式表述，则应给予高分。若只有部分一致或存在偏差，则根据一致的程度适当扣分。如果没有预测判决内容，扣10分。 2. \textbf{完整性}：评估预测答案是否涵盖了所有标准答案中提到的关键点，包括但不限于当事人、具体金额、责任判定、费用承担等。如果遗漏重要信息，则应相应扣分。 3. \textbf{准确性}：检查预测答案中提及的细节、数字、日期和法律依据是否与标准答案保持一致。任何错误信息均需扣分，并且严重错误应该导致更多的扣分。 4. \textbf{客观性与专业性}：预测答案应客观反映法案内容并使用恰当的法律术语。主观臆断或非专业表达需酌情扣分。 预测答案：\{prediction\} 参考答案：\{ground\_truth\} \\ \midrule
Translation    & You are an expert master in machine translation. Please score the predicted answer against the standard answer out of 10 points based on the following criteria: Content accuracy: Does the predicted answer accurately reflect the key points of the reference answer? Level of detail/completeness: Does the predicted answer cover all important points from the standard answer? Content redundancy: Is the predicted answer concise and consistent with the style of the standard answer? Respond following the format: "Content accuracy x points, level of detail/completeness x points, ..., total score: x points". The total score is the average of all the scores. Do not give reasons for your scores. Predicted answer: \{prediction\} Reference answer: \{ground\_truth\} \\ \bottomrule
\end{tabular}
\caption{Task instructions for model performance evaluation. The placeholder \{prediction\} and \{ground\_truth\} represent model prediction and reference answer, respectively.}
\label{tab:task_instructions}
\end{table*}
\end{CJK}




\section*{Appendix}
\section{Examples for Specialized Tasks}
\label{app:examples}

Table~\ref{tab:task_descriptions} presents task examples as prompts and corresponding reference responses for each specialized task, including intent recognition, text summarization, legal judgment prediction, and low-resource translation.

\section{Strategy for Grouping Experts}
\label{app:greedy}
To group experts together and simulate coarse-grained mixture-of-experts transformer models, we calculate expert similarity and group the experts by maximizing in-group similarities using a greedy search algorithm.

We sample data from the alignment dataset, containing 32 samples each with a sequence length of 4096, to calculate the similarity between experts. We initialize a co-occurrence matrix for all expert pairs as a zero matrix. For each pair of experts that occur simultaneously in a token's Top-6 expert choices, we increment their score by 1 in the matrix. After iterating through the dataset, we calculate the similarity between each pair of experts $i$ and expert $j$ using the cosine similarity between the vectors of row $i$ and row $j$ in the matrix. 

To obtain an expert grouping strategy through greedy search, we calculate the average intra-group similarity (the average pairwise similarity of all experts within the group) for all possible K-expert groups (where K is the group size, either 2 or 4) from the 64 non-shared experts out of the 66 experts in each layer. We then select the K-expert group with the highest score. For the remaining unselected experts, we repeat this process until all experts are selected and grouped.

\section{Analysis of Expert Affinity Sample Size}
\label{app:samplesize}
\begin{figure}[t]
    \centering
    \includegraphics[width=\linewidth]{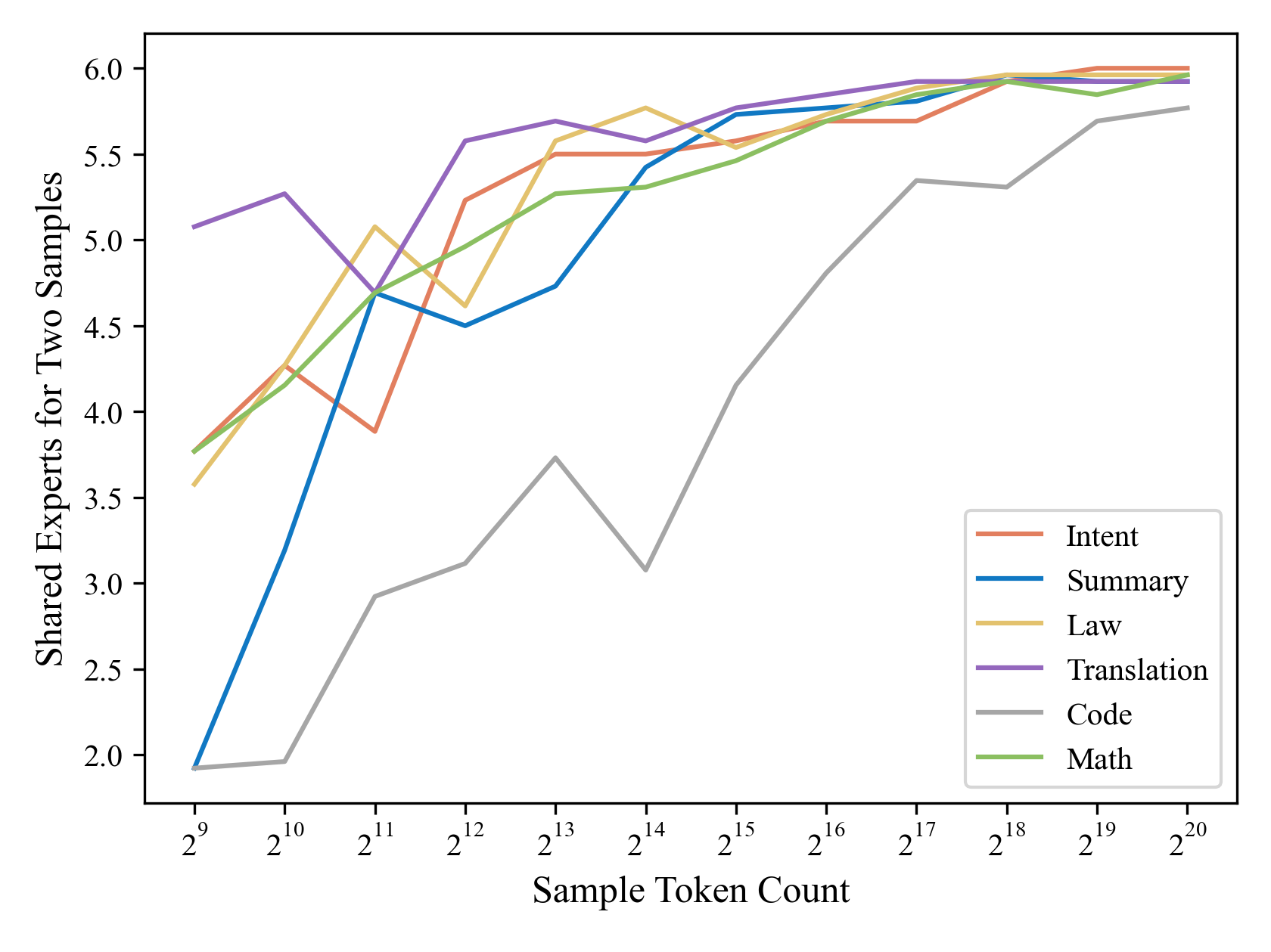}
    \caption{Results of the shared Top-6 routed experts in two independent samples of a task. The x-axis represents the sample size, and the y-axis shows the shared Top-6 routed experts averaged by model layers.}
    \label{fig:samplesize}
\end{figure}

To evaluate the amount of data needed to identify the most relevant experts for a task, we independently sample two sets of data from the training set for each of the six tasks and calculate the shared Top-6 experts between the two sets. The results are shown in Figure~\ref{fig:samplesize}. As the sample size reaches $2^{17}$ (i.e., 32 samples with a sequence length of 4096), all tasks exhibit a high number of shared experts between the two samples. This indicates that the sample size is sufficiently large to select the top-relevant experts for the tasks.

\section{Detailed Results for Ablations on Training Shared Parameters}
\label{app:ablation-shared}
We present two tables that summarize the performance of various methods with different configurations for training shared or non-shared parameters. Table~\ref{tab:general_sharednonshared} shows results on general tasks, and Table~\ref{tab:specialized_sharednonshared} focuses on specialized tasks. The results indicate that training only task-relevant non-shared experts consistently maintains the best general task performance. Additionally, training task-relevant non-shared experts and all shared parameters yields the best specialized task performance, short of full-parameter fine-tuning.

\afterpage{
\begin{figure*}[t]
    \centering

    \begin{subfigure}[b]{0.48\textwidth}
        \centering
        \includegraphics[width=\textwidth]{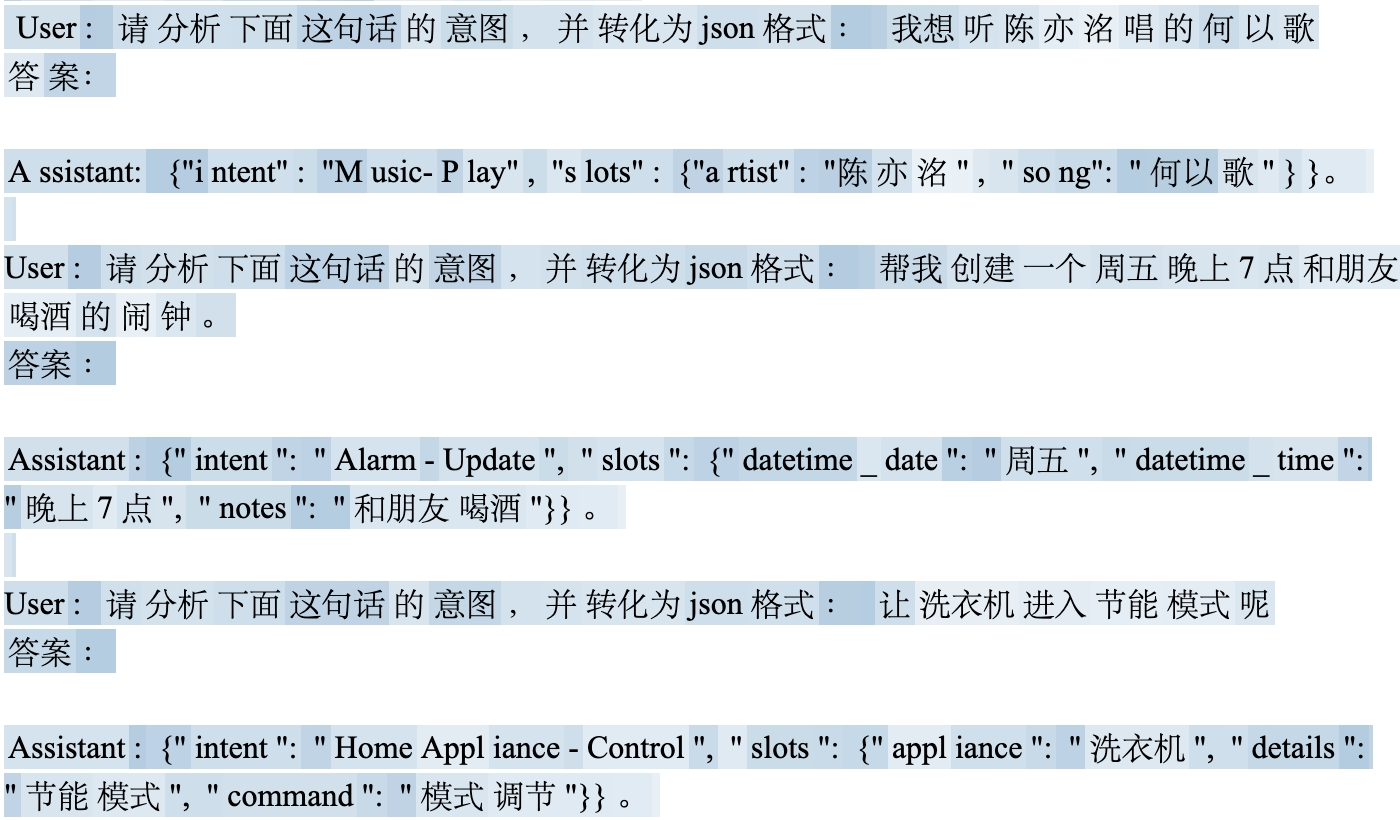}
        \caption{Intent recognition}
    \end{subfigure}
    \hfill
    \begin{subfigure}[b]{0.48\textwidth}
        \centering
        \includegraphics[width=\textwidth]{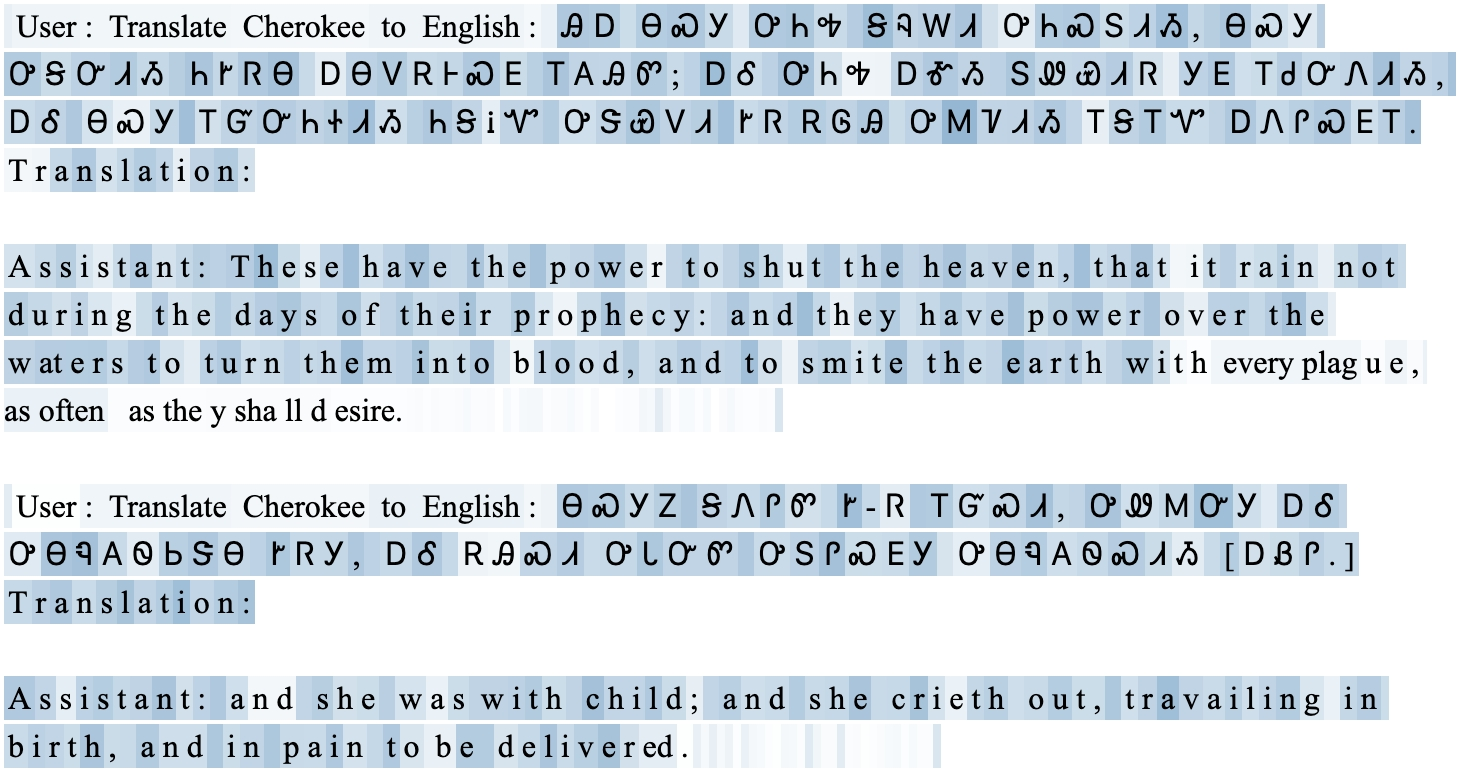}
        \caption{Low-resource translation}
    \end{subfigure}
    
    \begin{subfigure}[b]{0.48\textwidth}
        \centering
        \includegraphics[width=\textwidth]{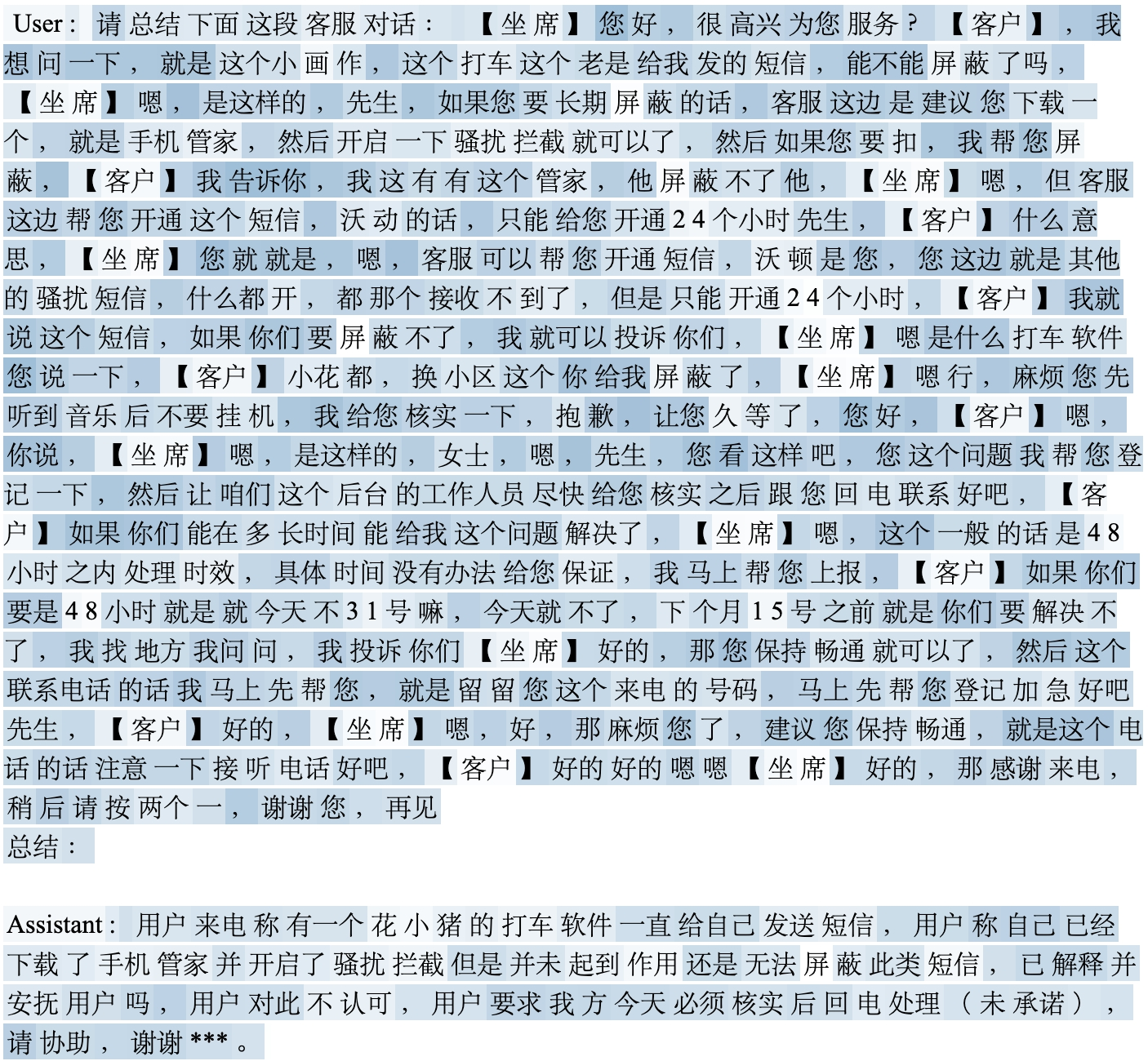}
        \caption{Text summarization}
    \end{subfigure}
    \hfill
    \begin{subfigure}[b]{0.48\textwidth}
        \centering
        \includegraphics[width=\textwidth]{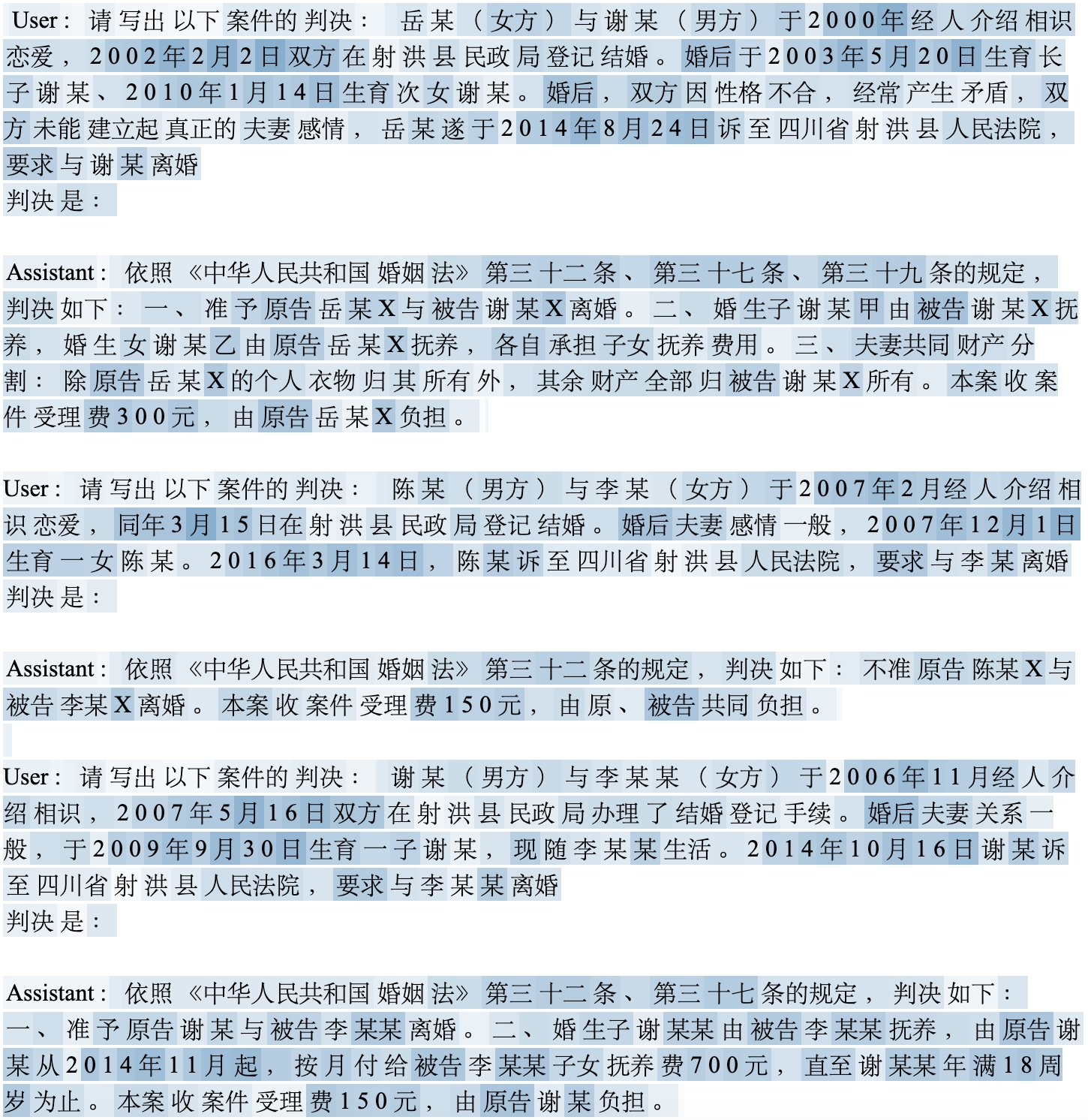}
        \caption{Legal judgment prediction}
    \end{subfigure}
    
    \begin{subfigure}[b]{0.48\textwidth}
        \centering
        \includegraphics[width=\textwidth]{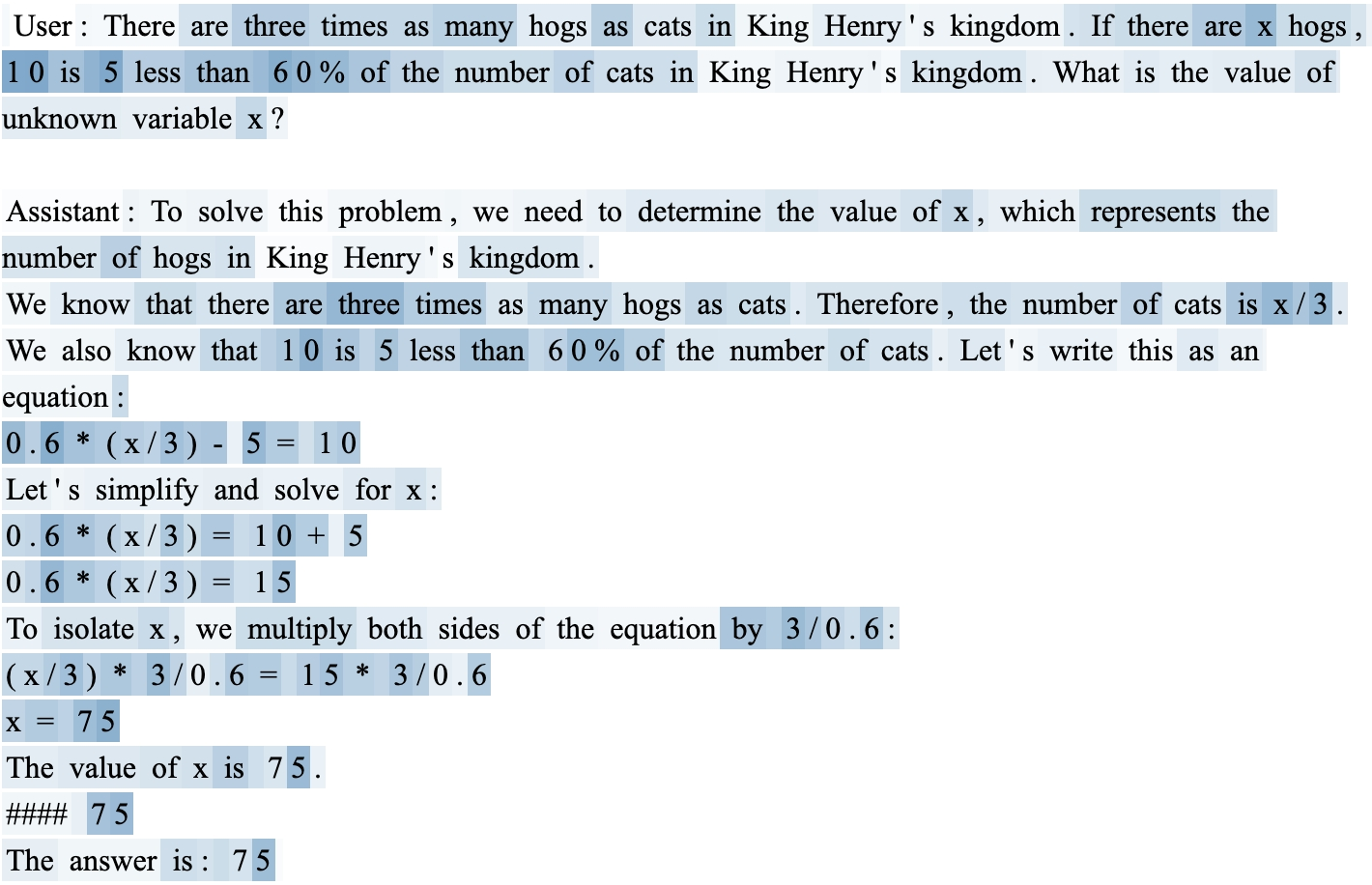}
        \caption{Math domain}
    \end{subfigure}
    \hfill
    \begin{subfigure}[b]{0.48\textwidth}
        \centering
        \includegraphics[width=\textwidth]{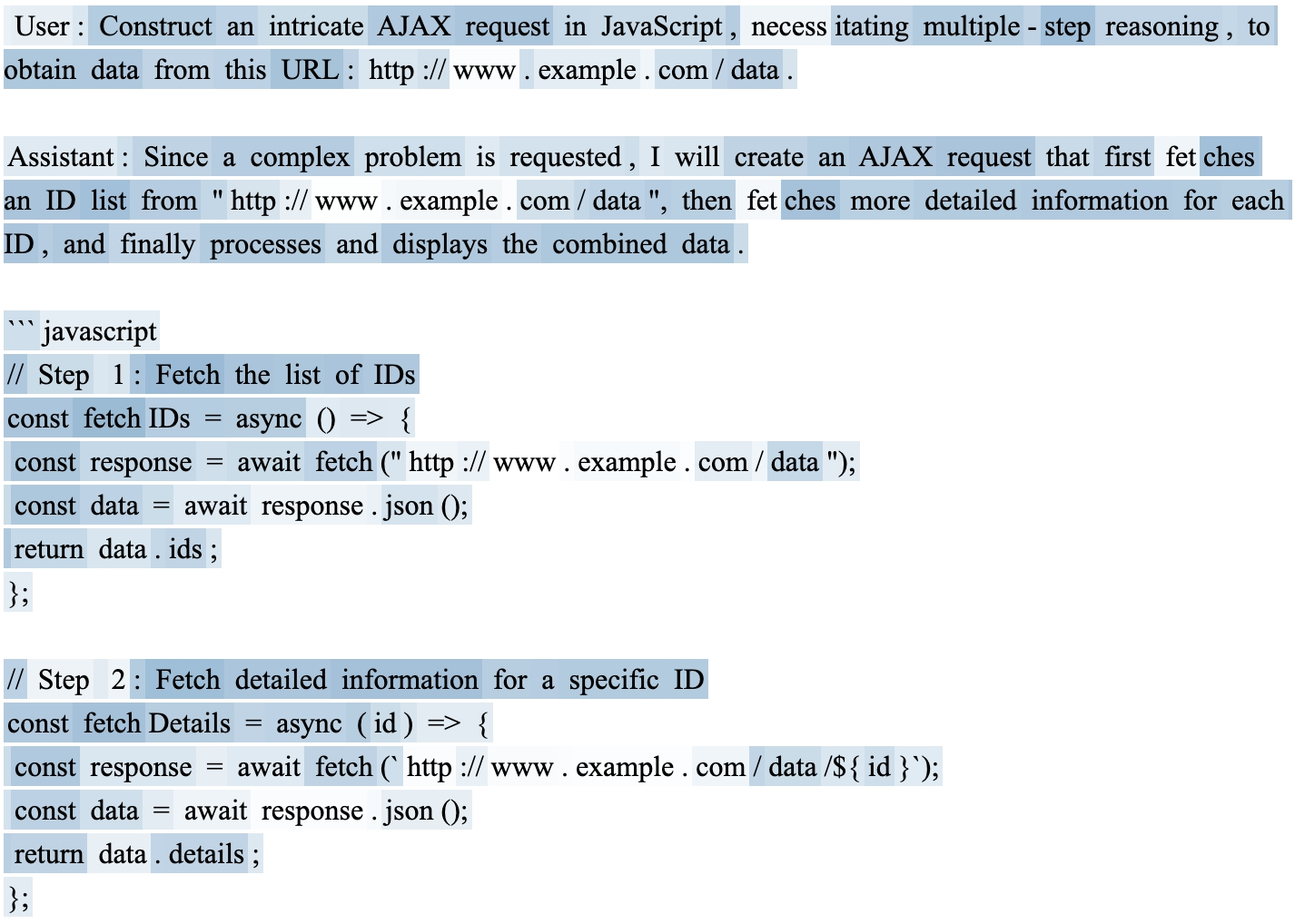}
        \caption{Code domain}
    \end{subfigure}
    \caption{Examples for our ESFT method showing the proportion of trainable routed experts among all tokens for each task. Deeper tokens indicate more trainable experts across all 26 layers (top-6 experts per layer). The parameter $p$ is set to 0.2 for the token selection ratio. Results show that our method, even handling only about 20\% of expert choices, covers a wide range of key task-relevant words.}
    \label{fig:qualitative}
\end{figure*}
}
\section{Qualitative Examples of the Expert Choices}
\label{app:qualitative}

We present qualitative examples of the amount that routed experts are trainable among all tokens for each task in Figure~\ref{fig:qualitative}. Each subfigure demonstrates examples drawn from a task. 
Deeper tokens indicate more trainable experts across all 26 layers (top-6 experts per layer). The parameter $p$ is set to 0.2 for the token selection ratio. Results show that our method, even handling only about 20\% of expert choices, covers a wide range of key task-relevant words.

\begin{CJK}{UTF8}{gbsn}
For example, in the Intent recognition task, the deepest tokens are ``意图'' (Intent); in the legal judgment task, the deepest tokens include ``婚后'' (Post-marriage), ``要求''(request), ``原告'' (plaintiff) and ``被告'' (defendant); in the Math task, the deepest tokens are mainly numerical tokens such as ``3'', ``5'', ``6'' and ``7''; in the Code task, the deepest tokens are key words like ``const'', or important commentary words like ``Fetch the list of IDs''.
\end{CJK}

\section{The Impact of Mixing Alignment Data for Training}
\label{app:sft-mix}

We adopt a 1:1 ratio for downstream task data and alignment data for all methods during training to better maintain general task performance. This manual ratio is kept constant to avoid the significant additional costs associated with fine-tuning the ratio for each task.

In this section, we present performance comparisons across various methods and tasks to reveal the impact of mixing alignment data during training. Table \ref{tab:mix-main} presents the performance on downstream specialized tasks, and Table \ref{tab:mix-general} shows the performance on general tasks.

The results indicate that FFT and LoRA benefit from the inclusion of alignment data, leading to improved performance in general tasks while only slightly decreasing performance in downstream tasks. Conversely, our ESFT method does not exhibit the same advantage. Specifically, mixing alignment data does not result in performance increases in either general or downstream tasks. The findings suggest that ESFT is inherently capable of adapting to downstream tasks without significant performance degradation in general tasks, even without added alignment data. This highlights the robustness and adaptability of ESFT in diverse task settings.

\section{Evaluation Instructions for Specialized Tasks}

Table~\ref{tab:task_instructions} presents the detailed criteria to evaluate specialized tasks including text summarization, legal judgment prediction, and low-resource translation.
Each task includes specific instructions on assessing predicted answers against reference answers, focusing on aspects such as content accuracy, completeness, relevance, and consistency. 

\label{sec:app-evaluation}

\section{Evaluating Math and Code as General Tasks}

We investigate the Math and Code performance of models trained on adaptation tasks (i.e., Intent, Summary, Law, Translation), as these domains reflect the model's general ability if not specifically trained on them. We report numbers with the setting of training on only downstream task data. Results in Table~\ref{tab:mathcode} show that FFT and LoRA would lead to significant performance drops in the Math and Code domain, having average performance drops of 9.0 and 12.4, respectively. Notably, our ESFT method retains performance significantly better compared to FFT and LoRA, with an average performance drop of less than 1.

\label{app:mathcode}

\end{document}